%% file: main.tex
\documentclass[10pt, svgnames]{article} 
\usepackage[preprint]{tmlr}

\input{math_commands.tex}

\usepackage{hyperref}
\usepackage{url}

\usepackage{xcolor}
\usepackage{graphicx}
\usepackage{cleveref}
\usepackage{hyperref}
\usepackage[framemethod=TikZ]{mdframed}
\usepackage{mathtools, amsthm}
\usepackage{comment}
\usepackage{wrapfig}
\usepackage{tabularx}
\usepackage{multicol}
\usepackage{booktabs}
\usepackage{pifont}
\usepackage{adjustbox}
\usepackage[numbers, sort&compress]{natbib}
\usepackage{tikz}
\usepackage{tikz-3dplot}
\usetikzlibrary{calc,math,backgrounds}
\usetikzlibrary{decorations}
\usetikzlibrary{decorations.markings}
\usetikzlibrary{decorations.pathreplacing,calligraphy}
\usetikzlibrary{shapes.geometric}

\usepackage{enumitem}

\newtheorem{definition}{Definition}

\newcommand{\dott}{\scriptsize\ding{108}}

\newcommand\independent{\protect\mathpalette{\protect\independenT}{\perp}}
\def\independenT#1#2{\mathrel{\rlap{$#1#2$}\mkern2mu{#1#2}}}

\pgfdeclarelayer{back}
\pgfdeclarelayer{front}
\pgfsetlayers{back,main,front}

\makeatletter
\pgfkeys{%
  /tikz/on layer/.code={
    \pgfonlayer{#1}\begingroup
    \aftergroup\endpgfonlayer
    \aftergroup\endgroup
  },
  /tikz/node on layer/.code={
    \gdef\node@@on@layer{%
      \setbox\tikz@tempbox=\hbox\bgroup\pgfonlayer{#1}\unhbox\tikz@tempbox\endpgfonlayer\egroup}
    \aftergroup\node@on@layer
  },
  /tikz/end node on layer/.code={
    \endpgfonlayer\endgroup\endgroup
  }
}

\def\node@on@layer{\aftergroup\node@@on@layer}

\makeatother

\hypersetup{%
    colorlinks = true,
    citecolor  = blue,
    linkcolor  = blue,
}

\newmdenv[ 
  linecolor=purple,
  linewidth=2pt,
  topline=false,
  bottomline=false,
  rightline=false,
  leftline=true,
  skipabove=\topsep,
  skipbelow=\topsep,
  backgroundcolor=purple!3
]{leftrule}

\newenvironment{element}[1]
    {
    \vspace{15pt}
    \begin{leftrule}[
        frametitle={\textcolor{purple}{CDL Element:} {\em #1}}
    ]}
    {
    \end{leftrule}
    \vspace{8pt}
    }

\tikzset{>=latex}

\title{Causal Deep Learning\\ \large \em Encouraging impact on real-world problems through causality}

\author{\name Jeroen Berrevoets \email jeroen.berrevoets@maths.cam.ac.uk \\
      \addr DAMTP, University of Cambridge
      \AND
      \name Krzysztof Kacprzyk \email kk751@cam.ac.uk \\
      \addr DAMTP, University of Cambridge
      \AND
      \name Zhaozhi Qian \email zhaozhi.qian@maths.cam.ac.uk\\
      \addr DAMTP, University of Cambridge
      \AND
      \name Mihaela van der Schaar \email mv472@cam.ac.uk\\
      \addr DAMTP, University of Cambridge\\
      The Alan Turing Institute}

\def\month{MM}  
\def\year{YYYY} 
\def\openreview{\url{https://openreview.net/forum?id=XXXX}} 

\begin{document}

\maketitle

\begin{abstract}

Causality has the potential to truly transform the way we solve a large number of real-world problems. 
Yet, so far, its potential largely remains to be unlocked as causality often requires crucial assumptions which cannot be tested in practice.  
To address this challenge, we propose a new way of thinking about causality-– we call this {\it causal deep learning}.  
Our causal deep learning framework spans three dimensions:  
(1) a structural dimension, which incorporates partial yet testable causal knowledge rather than assuming either complete or no causal knowledge among the variables of interest; 
(2) a parametric dimension, which encompasses parametric forms that capture the type of relationships among the variables of interest; 
 and  
(3) a temporal dimension, which captures exposure times or how the variables of interest interact (possibly causally) over time.
Causal deep learning enables us to make progress on a variety of real-world problems by leveraging partial causal knowledge (including independencies among variables) and quantitatively characterising causal relationships among variables of interest (possibly over time).
Our framework clearly identifies which assumptions are testable and which ones are not, such that the resulting solutions can be judiciously adopted in practice.
Using our formulation we can combine or chain together causal representations to solve specific problems without losing track of which assumptions are required to build these solutions, pushing real-world impact in healthcare, economics and business, environmental sciences and education, through causal deep learning.

\end{abstract}

\clearpage
\tableofcontents

\clearpage
\section{Introduction}

Causality holds the promise to transform the way we solve a large number of real-world problems \citep{pearl2018book,cml4impact}. 
Unlocking this promise amounts to adapting causality to the real world with its highly complex, unstructured, and abundance of data.
Beyond rich datasets, causal inference methods typically rely on diverse and extensive prior knowledge about the causal nature of the underlying system, many of which are not available or even testable in the application domain \citep{imbens2020potential}. 
This has led to two consequences. 
Firstly, practitioners may shy away from adopting causal inference in general because the available causal knowledge is not sufficient, let alone complete. 
Secondly, the emphasis on causal knowledge may overshadow other important considerations such as the statistical, functional, or temporal properties. 
These consequences may lead the practitioner to under-utilize these methods which ultimately results in sub-optimal solutions.

We want to encourage researchers and practitioners to take causality to its next step: real-world impact.
For this, we recognise the need to integrate causal inference with the sophisticated modelling capabilities of deep learning. 
This integration is particularly vital in real-world domains where understanding causal relationships can lead to better decision-making and predictions.

In this article, we introduce a new framework which enables pragmatically adopting causality ideas to solve real-world problems: Causal Deep Learning (CDL). 
Informally, CDL leverages {\it partial causal knowledge} (among some and not necessarily all variables of interest), and {\it quantitatively characterises the functional form} (among variables of interest) and the {\it time-evolution of the variables} to provide significant insights to researchers and decision-makers.
The reason why the above properties are important is two-fold:
(i) We need a good way to match a model with prior knowledge in a complex system, this should allow information of any type (be it no information, partial information, or full information); and
(ii) We need a good way to {\it evaluate} a solution.
This latter point is important in any real-world setting, especially settings where we rely on modelling solutions to support impactful decisions.
Inherently being built on a strong set of assumptions, model validation is a particularly tricky endeavour in causality and one which CDL can help us solve \citep{pearl2001bayesianism}.

{\bf Illustrative examples.}
Let us explore four diverse domains where causal deep learning, characterised by its focus on causal structure, functional relationships among variables of interest, and time, to understand the need for this new way of thinking.

First, consider the medical and healthcare domain, where determining the effectiveness of medications is paramount. 
Traditional models from pharmacology or physiology often struggle to capture the complexity of drug interactions and patient responses. 
To make progress in studying the effects of medications on health outcomes, CDL might focus on the causal relationship between the drug dosage and patient recovery rate while controlling for other variables like age, gender, and pre-existing conditions. 
In such a case, CDL would not assert complete causality of overall health factors but rather isolate the drug’s impact. 
By employing only a partial causal structure, CDL can model the nuanced relationship between drug dosage and patient recovery, considering both linear and non-linear effects. 
Moreover, by incorporating temporal dynamics, CDL may unravel how medication effectiveness evolves over time, providing critical insights for personalized medicine.

In economics and business, the relationship between interest rate changes and consumer spending is a classic example of a complex causal interaction. 
Traditional econometric models might fail to capture the intricacy of this relationship. 
CDL, with its ability to handle non-linear and high-dimensional data, can provide a more robust understanding. 
By adopting a pragmatic approach, CDL could enable the investigation of partial causality between interest rate changes and consumer spending. 
While acknowledging that other factors like education, employment rates, inflation, and economic policies also impact spending, the focus here is on understanding how variations in interest rates specifically influence consumer behaviour. 
By considering the temporal dimension, CDL may also uncover lag effects, where changes in interest rates take time to manifest in consumer behaviour, a crucial insight for policymakers.

Environmental science, especially the study of climate change, presents another compelling case for causal deep learning. 
An interesting research aspect might be to explore the causal relationship between carbon dioxide emissions and global temperature increase, acknowledging that other factors like deforestation and solar radiation also play roles in climate change. 
Using such a pragmatic CDL approach which requires only partial causal structure would allow researchers to isolate and understand the specific impact of CO2 emissions. 
In addition, the relationship between CO2 emissions and global temperature increase is not straightforward and likely non-linear. 
Here, causal deep learning can capture the complexity of this relationship beyond traditional linear models. 
Additionally, understanding how this relationship evolves over time is critical for predicting future climate patterns and for formulating effective environmental policies.

Lastly, in the education sector, the impact of classroom size on student performance is a topic of ongoing debate. 
While simple linear models might suggest a straightforward inverse relationship, the reality is likely more nuanced. 
Causal deep learning can help in identifying not just whether, but how class size impacts student performance, including potential threshold effects or non-linear dynamics. 
Furthermore, by examining how this relationship changes over an academic year or across different educational stages, deeper insights can be gained into effective educational planning and resource allocation.

In these and other real-world domains, causal deep learning can be used as a powerful framework to make progress. 
Unlike current causal models in machine learning, CDL allows for a nuanced understanding of causal relationships without requiring complete causal knowledge, learns in a data-driven manner the parametric form of the relationships among the variables of interest using powerful deep-learning models, and importantly, considers the dimension of time. 
This approach can uncover insights that traditional models might miss, leading to more effective interventions and policies across various sectors. The exploration of causal deep learning in these diverse real-world domains not only underscores its versatility but also highlights its potential to revolutionize how we understand and interact with the world around us.

With our detailed definitions, we can better align a problem with the (causal) solution.
While research in causality has a long history of listing out the various assumptions necessary to identify causal effects, these assumptions often do not map easily into practice.
Furthermore, deep learning has no such history, typically because most deep learning architectures are assumed to be non-parametric.
However, not all problems require the enormous flexibility of deep neural networks, but may still be non-linear.
Our proposal encompasses all the above in hopes of aiding the correct and useful adoption of causal deep learning methods.

\section{CDL in 3 dimensions: Structural, Parametric, and Temporal} \label{sec:cartography}

CDL aims to provide practical guidelines that yield more informed and robust models to solve real-world problems.
We introduce these guidelines under two main axes: (i) {\it the structural scale} concerning the links between variables, described in \cref{sec:scale}; and (ii) {\it the parametric scale} concerning the shape of the functions that model links between variables, described in \cref{sec:param}.
These two axes describe the first two (of three) directions in which CDL is best described. 
The third is time, which due to its special meaning in causality (causes precede effects), is worthy of a completely separate discussion.
As we will show, CDL's guidelines help us to combine models, build pipelines, and test settings of the data-generating process to avoid unexpected bias and errors.

\subsection{Preamble: distribution factorization} \label{sec:factor}

Before explaining the structural and parametric scale, we revisit the fundamental concept of distribution factorization, which states how a distribution can be factorised as a series of functions.
\begin{definition}[factor]
Let $\mathbf{X}$ be a set of random variables. We define a factor $f$ to be a function from $\text{Val}(\mathbf{X}) \to \mathbb{R}$. A factor is non-negative if all its entries are non-negative. The set of variables $\mathbf{X}$ is called the scope of the factor and denoted $\text{Scope}[f]$ \citep[Definition~4.1]{koller2009probabilistic}.
\end{definition}
In its most general form, a joint distribution admits a factorisation as,
\begin{equation}\label{eq:factor}
    p(\mathbf{X}) = \frac{1}{Z} \prod_{i\in[I]} f_i(\mathbf{X}^{(i)}),
\end{equation}
where $Z$ is a normalising constant, and one factor $f_i$ takes a set of variables, $\mathbf{X}^{(i)} \subseteq \mathbf{X}$, as arguments \citep{koller2009probabilistic}. 

Notice that in \cref{eq:factor} we iterate over $I$ factors. However, the number of factors and each $\mathbf{X}^{(i)}$ is, at this stage, not yet fully determined. A trivial example of such a factorisation is the decomposition of a joint distribution into conditional distributions: $p(A, B) = f_1(A, B)f_2(B)$ with $f_1(A, B) = p(A | B)$ and $f_2 = p(B)$. As another trivial example, one could have a factorisation which is comprised of the least amount of factors: $p(\mathbf{X}) = \frac{1}{Z} f(\mathbf{X})$.

Factorization plays a central role in CDL because it captures both the structural and parametric properties of the underlying system. The structural properties manifest in the scope of each factor $\mathbf{X}^{(i)}$ whereas the parametric properties manifest in the functional form of $f$. Although distribution factorization is originally a statistical concept, it also has a close connection with various causality frameworks, such as structural causal models. Therefore, we use it as a starting point to introduce the scales.

\subsection{The structural scale and rung 1.5} \label{sec:scale}

Crucial to causal deep learning is the ability to leverage knowledge about the relationships between variables, i.e., which variables are placed in the same factor $f$.  
Conveniently depicted as graphs, we can order these relations on a scale from less detailed to more detailed -- this gives the structural scale, which is central to our map of causal deep learning.

\begin{figure}[t]
    \centering
    \begin{tikzpicture}[
        dot/.style={circle, fill=black, draw=black,  minimum size=2mm, inner sep=0},
    ]
    
    \node[inner sep=0] (start) at (0, 0) {};
    \node[inner sep=0] (end) at (8, 0) {};
    
    \node[dot, inner sep=0] (blob) at (start) {};
    \node[anchor=south, inner sep=0] (unknown_node) at ($(start) - (0, .5)$) {\textcolor{purple}{unknown}};
    \node[rounded corners, draw=none, fill=purple!5] at ($(unknown_node) - (0, .5)$) {\textcolor{blue}{No tests needed}};
    \begin{scope}[shift={($(blob) + (0,.5)$)}]
        \node[inner sep=1] (S) at (-.8, 0) {$S$};
        \node[inner sep=1] (C) at (0, 0) {$C$};
        \node[inner sep=1] (D) at (.8, 0) {$D$};
        
        \draw (S) -- (C);
        \draw (C) -- (D);
        \draw (D) to [out=135, in=45] (S);
    \end{scope}

    \node[dot, inner sep=0] (directed) at ($(start)!.5!(end)$) {};
    \node[anchor=south, inner sep=0] (plausible_node) at ($(start)!.5!(end) - (0, .565)$) {\textcolor{purple}{plausible}};
    \node[rounded corners, draw=none, fill=purple!5] at ($(plausible_node) - (0, .5)$) {\textcolor{blue}{Can be tested!}};
    \begin{scope}[shift={($(directed)+(0,.5)$)}]
        \node[inner sep=1] (S) at (-.8, 0) {$S$};
        \node[inner sep=1] (C) at (0, 0) {$C$};
        \node[inner sep=1] (D) at (.8, 0) {$D$};
        
        \draw[->, densely dotted] (C) to [in=45, out=135] (S);
        \draw[->, densely dotted] (C) to [out=45, in=135] (D);
        
        \draw[->] (S) -- (C);
        \draw[->] (C) -- (D);
        
        \draw[->, densely dashed] (D) to [out=225, in=315] (C);
        \draw[->, densely dashed] (C) to [out=225, in=315] (S);
    \end{scope}
    
    \node[dot, inner sep=0] (causal) at (end) {};
    \node[anchor=south, inner sep=0] (causal_node) at ($(end) - (0, .5)$) {\textcolor{purple}{causal}};
    \node[rounded corners, draw=none, fill=purple!5] at ($(causal_node) - (0, .5)$) {\textcolor{blue}{Untestable}};
    
    \begin{scope}[shift={($(causal)+(0,.5)$)}]
        \node[inner sep=1] (S) at (-.8, 0) {$S$};
        \node[inner sep=1] (C) at (0, 0) {$C$};
        \node[inner sep=1] (D) at (.8, 0) {$D$};

        \draw[->] (S) -- (C);
        \draw[->] (C) -- (D);
    \end{scope}
    
    \node[blue] at ($(start) + (0, 2.25)$) {{\bf less} structure};
    \node[blue] at ($(end) + (0, 2.25)$) {{\bf lots of} structure};
    
    \draw[very thick] (start) -- (end);

    \draw[thick, blue, latex-latex, rounded corners=3mm] ($(start)+(0, 1.15)$) -- ($(start) + (0, 1.65)$) -- ($(end) + (0, 1.65)$) -- ($(end) + (0,1.15)$);

    \end{tikzpicture}
    \caption{{\bf The structural scale.} The first axis on the map of causal deep learning is the structural scale which categorises different structure types. The two extremes are non-informative structures or a complete causal graph. Note that the structural scale makes no parametric assumptions on these structures (parametric assumptions are the focus of the parametric scale in \cref{fig:param}).}
    \label{fig:scale}
    \rule{\textwidth}{.5pt}
\end{figure}
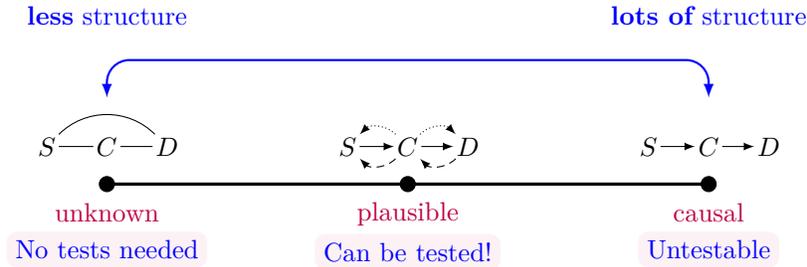

\cref{fig:scale} depicts the structural scale, which ranges from \textcolor{blue}{{\bf less} structure}, to \textcolor{blue}{{\bf more} structure}. The structural scale governs information about the statistical or causal relationships in a system. Essentially, it quantifies how much we know, expressed in graphical structures. Below, we discuss each level using an illustrative example of three variables: smoking, denoted as $S$; cancer, denoted as $C$; and death, denoted as $D$.

\textbf{\em Level 1 -- no structure.} At the lowest end of our scale we have structures that exhibit zero knowledge, i.e., $p(\mathbf{X}) = \frac{1}{Z} f(\mathbf{X})$. These structures connect all variables because all variables appear in the same factor. An example of such a structure is denoted as, 
\begin{equation} \label{eq:blob}
    \begin{tikzpicture}[
        baseline=-1.1mm
    ]
        \node[inner sep=1] (S) at (-.8, 0) {$S$};
        \node[inner sep=1] (C) at (0, 0) {$C$};
        \node[inner sep=1] (D) at (.8, 0) {$D$};
        
        \draw (S) -- (C);
        \draw (C) -- (D);
        \draw (D) to [out=135, in=45] (S);
        
        \node at (4, 0) {\textcolor{purple}{\bf (unknown).}};
    \end{tikzpicture}
\end{equation}
While technically a structure, we have no idea about the potential relationships these variables have concerning each other. Having a fully connected structure such as in \cref{eq:blob} means that we assume {\it potential} dependence between each variable. Assuming a fully connected structure is not a strict assumption. In our example above, the structure in \cref{eq:blob} states that ``smoking, cancer and death'' are possibly related, which is a very general statement indeed.

That is because relatedness is a flexible term. While having an edge may appear to be a strict statement, the edge does not express the {\it level} at which two variables are related. In fact, the {\it absence} of an edge is a much stronger assumption which we discuss next.

\textbf{\em Level 2 -- plausibly causal structures.} Let us assume now that we know a little more: cancer is the reason why smoking is related to death. Statistically, we can express this assumption as $S \independent D | C$. In Level 2, we assume {\it statistical independence}. Independence is not directly the same as assuming some structural information, however, it can be expressed as such.

The easiest thing we can do to reflect $S \independent D | C$ is to explicitly remove the edge \begin{tikzpicture}[
    baseline=-1.1mm
]
    \node[inner sep=1] (S) at (-.8, 0) {$S$};
    \node[inner sep=1] (D) at (.8, 0) {$D$};

    \draw (D) to [out=135, in=45] (S);
\end{tikzpicture} from \cref{eq:blob} as we have below in \cref{eq:association},
\begin{equation} \label{eq:association}
    \begin{tikzpicture}[
        baseline=-1.1mm
    ]
        \node[inner sep=1] (S) at (-.8, 0) {$S$};
        \node[inner sep=1] (C) at (0, 0) {$C$};
        \node[inner sep=1] (D) at (.8, 0) {$D$};
        
        \draw (S) -- (C) -- (D);
        
    \end{tikzpicture}.
\end{equation}

Crucially, however, there exist more structures which reflect $S \independent D | C$. Consider the three structures below,

\begin{minipage}{\textwidth}
\vspace{5pt}
    \begin{minipage}[t]{.32\textwidth}
        \centering
        \begin{tikzpicture}
            \node[inner sep=1] (S) at (-.8, 0) {$S$};
            \node[inner sep=1] (C) at (0, 0) {$C$};
            \node[inner sep=1] (D) at (.8, 0) {$D$};
            
            \draw[->] (S) -- (C);
            \draw[->] (C) -- (D);
        \end{tikzpicture},
    \end{minipage}
    ~
    \begin{minipage}[t]{.32\textwidth}
        \centering
        \begin{tikzpicture}
            \node[inner sep=1] (S) at (-.8, 0) {$S$};
            \node[inner sep=1] (C) at (0, 0) {$C$};
            \node[inner sep=1] (D) at (.8, 0) {$D$};
            
            \draw[<-] (S) -- (C);
            \draw[->] (C) -- (D);
        \end{tikzpicture},
    \end{minipage}
    ~
    \begin{minipage}[t]{.32\textwidth}
        \centering
        \begin{tikzpicture}
            \node[inner sep=1] (S) at (-.8, 0) {$S$};
            \node[inner sep=1] (C) at (0, 0) {$C$};
            \node[inner sep=1] (D) at (.8, 0) {$D$};
            
            \draw[<-] (S) -- (C);
            \draw[<-] (C) -- (D);
        \end{tikzpicture},
    \end{minipage}
\vspace{1pt}
\end{minipage}
which all model the same independencies.

For the sake of our discussion, let us assume $S \independent D | C$ is correct\footnote{If cancer renders smoking and death independent, we imply that cancer is the {\it only} path from smoking to death, which is known to be false \citep{carter2015}. For example, smoking may result in chronic obstructive pulmonary disease (COPD) or ischemic heart disease, which can also lead to premature death. However, for illustration, we assume it is true.}. As (in)dependence is a statement of statistical association, we cannot assume that the structures based on it are causal. For example, \begin{tikzpicture}[
    baseline=-1.1mm
]
    \node[inner sep=1] (S) at (-.8, 0) {$S$};
    \node[inner sep=1] (C) at (0, 0) {$C$};
    \node[inner sep=1] (D) at (.8, 0) {$D$};
    
    \draw[<-] (S) -- (C);
    \draw[<-] (C) -- (D);
\end{tikzpicture}, implies that dying impacts whether or not the deceased is a smoker, which cannot be true.

The type of structure of interest in Level 2, are probabilistic graphical models (PGMs) \citep{koller2009probabilistic}. These include structures such as directed acyclic graphs (DAGs), undirected graphs, or even factor graphs. Depending on the set of independence statements one wishes to assume, one may express them as a certain type of PGM. 

For example, if one wishes to assume $S \independent D$ while also assuming $S \not\independent D | C$, we best use a DAG:
\begin{equation} \label{eq:collider}
    \begin{tikzpicture}
        \node[inner sep=1] (S) at (-.8, 0) {$S$};
        \node[inner sep=1] (C) at (0, 0) {$C$};
        \node[inner sep=1] (D) at (.8, 0) {$D$};
        
        \draw[->] (S) -- (C);
        \draw[<-] (C) -- (D);
    \end{tikzpicture},
\end{equation}
which is often called a ``collider'' structure. This collider structure (and more importantly the (in)dependence statements it implies) can only be modelled with a DAG.

Being able to model independence is a clear step up from Level 1, where we assumed nothing at all. While independence is a statement of statistical association (i.e. it is bidirectional and not causal), they do pose constraints for a later causal model. As we can check (in)dependence in data, a recovered causal model should respect (or even explain) these (in)dependence statements. Hence, the above structures are plausibly causal.

In level 2 we have no consensus on one graph in particular, only about dependence and independence (predominately statistical concepts). To illustrate Level 2 on our structural scale in \cref{fig:scale} above, we denote these plausible structures as,
\begin{equation} \label{eq:mec}
\begin{aligned}
    \begin{tikzpicture}[
        baseline=-1.1mm
    ]
        \node[inner sep=1] (S) at (-.8, 0) {$S$};
        \node[inner sep=1] (C) at (0, 0) {$C$};
        \node[inner sep=1] (D) at (.8, 0) {$D$};
        
        \draw[->, densely dotted] (C) to [in=45, out=135] (S);
        \draw[->, densely dotted] (C) to [out=45, in=135] (D);
        
        \draw[->] (S) -- (C);
        \draw[->] (C) -- (D);
        
        \draw[->, densely dashed] (D) to [out=225, in=315] (C);
        \draw[->, densely dashed] (C) to [out=225, in=315] (S);
        
        \node at (4, 0) {\textcolor{purple}{\bf (plausible).}};
    \end{tikzpicture}
    \end{aligned}
\end{equation}
where we combined all possible DAGs that respect, $S \independent D | C$ \citep{sun2006causal}. Applying the rules of d-separation \citep{verma1990}.

While Level 2 includes all types of PGMs, we chose to illustrate Level 2 using DAGs only, as DAGs are the structure of choice when modelling causality \citep{pearl2009causality}. In \cref{eq:mec} all of the structures are DAGs, but only one of them is causal (likely to be the first one).

\textbf{\em Level 3 -- causal structure.} That leaves assuming the exact causal structure as opposed to the many plausible causal structures in Level 2. With \cref{eq:causation} we say that smoking {\it causes} cancer which causes death. Such knowledge translates into the directed structure,
\begin{equation} \label{eq:causation}
    \begin{tikzpicture}[
        baseline=-1.1mm
    ]
        \node[inner sep=1] (S) at (-.8, 0) {$S$};
        \node[inner sep=1] (C) at (0, 0) {$C$};
        \node[inner sep=1] (D) at (.8, 0) {$D$};

        \draw[->] (S) -- (C);
        \draw[->] (C) -- (D);
        
        \node at (4, 0) {\textcolor{purple}{\bf (causal).}};
    \end{tikzpicture}
\end{equation}
Which was indeed one of the plausible DAGs in Level 2 (\cref{eq:mec}).

Comparing \cref{eq:association} with \cref{eq:causation} may lead to believe there is little difference between either end of the structural scale-- {\it nothing could be further from the truth \citep{pearl2018book,bareinboim2022pearl}!} Rather, the step from association to causation is very large and is a central argument in causality. So large in fact that we recognise several intermediary structures.

An important realisation is that each level of information further specifies the environment, i.e. with information, we {\it restrict} possible interpretations and data distributions. For example, from \cref{eq:blob,eq:association} we expect $S$ to change if $C$ is changed as both variables are associated with each other. However, \cref{eq:causation} tells us this is not necessarily the case: patients do not automatically start smoking when diagnosed with cancer. However, the reverse {\it is} true; when a cancer-patient {\it stops} smoking they have a higher chance of remission \citep{sheikh2021postdiagnosis}.

{\bf ``Rung 1.5''} Pearl's ladder of causation \citep{pearl2018book,bareinboim2022pearl} ranks structures in a similar way as we do, i.e., increasing a model's causal knowledge will yield a higher place upon his ladder. Like Pearl, we have three different levels in our scale. However, they do not correspond one-to-one. 

In particular, we find that our Level 2 is not well represented in Pearl's ladder of causation\footnote{Similarly, his counterfactual rung is not represented in our structural scale.}. We reason that our Level 2 {\it does} encode some prior knowledge into a model, which is more than encoding no prior knowledge at all. Yet, both would be categorised under Pearl's first rung. Of course, we recognise that the models categorised under Level 2 have no embedded {\it causal} knowledge. While Level 2 does not reach the level of a rung 2 model (but Level 3 does), we consider a Level 2 model to correspond with a hypothetical ``rung 1.5''.

At the structural level, we find that Pearl's rung 3 (counterfactuals) is the same as rung 2 (interventional). Of course, rung 3 goes {\it beyond} structure, in that it also requires functional knowledge. Instead, we dedicate an entirely new scale to functional knowledge as (like structure) we find there are different levels of functional knowledge one can assume (or learn). We introduce this additional scale in the following subsection.

\subsection{The parametric scale} \label{sec:param}

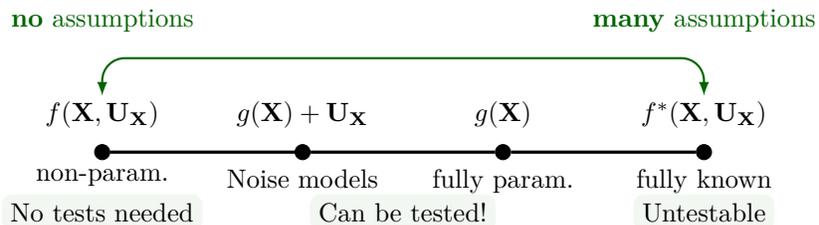
\begin{figure}[t]
    \centering
    \begin{tikzpicture}[
        dot/.style={circle, fill=black, draw=black,  minimum size=2mm, inner sep=0},
    ]
    
    \node[dot, label={below:non-param.}] (0moment) at (0, 0) {};
    \begin{scope}[shift={($(0moment) + (0,.5)$)}]
        \node[inner sep=1] (nonparam_node) at (0, 0) {$f(\mathbf{X}, \mathbf{U}_\mathbf{X})$};     
    \end{scope}
    
    \node[dot, label={below:fully known}] (multmoment) at (8, 0) {};
    \begin{scope}[shift={($(multmoment) + (0,.5)$)}]
        \node[inner sep=1] at (0, 0) {$f^*(\mathbf{X}, \mathbf{U}_\mathbf{X})$};
    \end{scope}
    
    \node[dot, label={below:Noise models}] (1moment) at ($(0moment)!.333!(multmoment)$) {};
    \begin{scope}[shift={($(0moment)!.333!(multmoment) + (0,.5)$)}]
        \node[inner sep=1] at (0, 0) {$g(\mathbf{X}) +  \mathbf{U}_\mathbf{X}$};
    \end{scope}

    \node[dot, label={below:fully param.}] (2moment) at ($(0moment)!.666!(multmoment)$) {};
    \begin{scope}[shift={($(0moment)!.666!(multmoment) + (0,.5)$)}]
        \node[inner sep=1] at (0, 0) {$g(\mathbf{X})$};
    \end{scope}

    \node[rounded corners, draw=none, fill=DarkGreen!5] at ($(0moment) - (0, .8)$) {\textcolor{black}{No tests needed}};
    \node[rounded corners, draw=none, fill=DarkGreen!5] at ($(1moment)!.5!(2moment) - (0, .8)$) {\textcolor{black}{Can be tested!}};
    \node[rounded corners, draw=none, fill=DarkGreen!5] at ($(multmoment) - (0, .8)$) {\textcolor{black}{Untestable}};

    \draw[very thick] (0moment) -- (1moment) -- (2moment) -- (multmoment);

    \node[DarkGreen] at ($(0moment) + (0, 1.75)$) {{\bf no} assumptions};
    \node[DarkGreen] at ($(multmoment) + (0, 1.75)$) {{\bf many} assumptions};
    \draw[thick, DarkGreen, latex-latex, rounded corners=3mm] ($(0moment)+(0, .75)$) -- ($(0moment) + (0, 1.25)$) -- ($(multmoment) + (0, 1.25)$) -- ($(multmoment) + (0,.75)$);

    \end{tikzpicture}
    \caption{{\bf The parametric scale.} A second axis in the map of causal deep learning. The parametric scale logs the type of assumptions made on the factors of the assumed distribution or model. The extremes are no assumptions at all (leaving completely non-parametric factors), or a fully known model. The parametric scale further discerns assumptions on the way noise interacts in the system ($\eps$) and the functional shape of the system's factorisation.}
    \label{fig:param}
    \rule{\textwidth}{.5pt}
\end{figure}

Our goal in this section is to organise the types of assumptions one can make at the parametric level of these factors, contrasting the structural assumptions one makes about dependence (as discussed previously in section \cref{sec:scale}). 
An important differentiating property of these assumptions is: the strictness expressed in the size of the accompanying hypothesis class \citep{vapnik2015uniform}, and whether or not one needs to test their validity {\it outside} what is possible given the available observational data.
We summarise the assumption types in \cref{fig:param} and introduce each separate component below.

\textbf{\em Level 1 -- non-parametric.} Level 1 of the parametric scale encompasses the least strict assumption: namely, there exists a factorisation such as \cref{eq:factor}. Beyond that, level 1 assumes no specific functional form of its factorisations, $f_i$.

Note here that assumptions such as iid data or value assumptions (such as numerical or ordinal variables) fall outside the scope of what is understood as a parametric scale above.

\textbf{\em Level 2 -- noise models.} The next level in our parametric scale is to first make assumptions on the composition of the set of random variables, $\mathbf{X}$. In particular, we assume that $\mathbf{X}$ can be separated in noise terms or exogenous variables $\mathbf{U}$ and variables $\mathbf{X}\setminus\mathbf{U}$ where each element in $\mathbf{U}$ corresponds with exactly one element in $\mathbf{X}\setminus\mathbf{U}$ and $|\mathbf{X}| = 2|\mathbf{U}|$.

The above assumption has implications for the factors. Specifically, if a factor takes $X \in \mathbf{X}\setminus\mathbf{U}$ as argument, it will always take $U_X \in \mathbf{U}$ (which is the corresponding noise variable of $X$) as argument also. Note that this separation alone is a {\it structural} assumption \citep{pearl2009causality}.

As a parametric scale, the Level 2 further states {\it how} the noise variables $\mathbf{U}_{\mathbf{X}^{(i)}}$ are incorporated in $f_i$ \citep{zhang2012identifiability,shimizu2011directlingam,hoyer2008nonlinear}.
In the case of causal structural equation models (where a random variable results from a function of that variable's causal parents), we have a common example in additive noise:
\begin{equation} \label{eq:add}
    \mathbf{X} = g(\textrm{Pa}(\mathbf{X})) + \mathbf{U}_\mathbf{X}.
\end{equation}

In short, \cref{eq:add} assumes we can decompose a factor into the noise, $\mathbf{U}_\mathbf{X}$, and some (non-parametric) function, $g$. While how $\mathbf{U}_\mathbf{X}$ interacts in $f$ is fixed, $g$ is still non-parametric and therefore only level 2 of this scale. 

Many prior works in causal discovery make assumptions of this level\citep{peters2017elements}. For example, in \citet{hoyer2008nonlinear} $g$ remains non-linear (through the use of neural networks) but $\mathbf{U}_\mathbf{X}$ is assumed additive independent (as above) resulting in a clear level 2 assumption.

Assumptions on the noise distribution (Gaussian, Uniform), which fall under the umbrella term of parametric assumptions, {\it can} be tested using statistical tests.
These types of distributional assumptions can, for example, be tested using the classical (non-parametric) Kolmogorov-Smirnov (K-S) test \citep{naaman2021tight}, or a Jarque-Bera test for normality \citep{jarque1980efficient}.

\textbf{\em Level 3 -- fully parametric.} Naturally, the next level is to make a parametric assumption on $g$ in \cref{eq:add}. These assumptions vary from making linear assumptions, to $n$ times differentiable, to many more types of assumptions on the functional form of $g$ and consequentially $f$. Note that because we make an assumption on $g$ (and not $f$ directly), we automatically assume that $f$ can be decomposed in $g$ and noise, i.e. Level 3 subsumes assumptions from Level 2.

Less parametric assumptions imply a larger model hypothesis class. For example, having linear assumptions allows much fewer possible factors than a non-linear class of assumptions. It has been shown that neural networks\footnote{In particular with a ReLU activation.} have a connection to non-parametric regression, leading to high-performance models \citep{schmidt2020nonparametric}.
As such, making a parametric assumption is typically done in favour of easier regression, but at the cost of less general model classes.

The testability of this type of parametric assumption largely depends on which assumption we are making. 
Testing based on, for example, a Savitzky-Golay filter \citep{savitzky1964smoothing} can be effective in testing a certain type of parametric assumption (such as a linear assumption).
In the specific case of linearity, the largest CUSUM test \citep{page1954continuous} in combination with a K-S test can be used for (in)validation.

\textbf{\em Level 4 -- fully known factors.} Rather than making assumptions on the functional class of the factors, some methods assume full access to a known set of factors which we term $f^*(\mathbf{X}, \mathbf{U}_\mathbf{X})$. This is of course a very strict assumption, resulting in the final level of the parametric scale.

To validate this type of function, one cannot rely on a standardised statistical test using only observational data \citep{pearl2001bayesianism}. 
As is the case in fields such as theoretical physics, validation of a specific function needs to happen experimentally (in a clinical trial, or by an experimental physicist).

With level 4 we conclude our parametric axis. This parametric axis has been the main focus of investigation of deep machine learning and statistics. The parametric scale is also important in causality (specifically econometric approaches to causality \citep{imbens2020potential}).

\textbf{Significance of the parametric scale}. We illustrate the benefit of the parametric scale with a concrete example---causal structure discovery models, which attempt to uncover causal graphs from observational data. Previously, {\it all causal structure discovery methods are categorised the same in the ladder of causation \citep{bareinboim2022pearl}}. By introducing a parametric scale, CDL allows for better differentiation between methods. For example, comparing \citet{zheng2018dags} with the later \citet{zheng2020learning} using the latter of causation, implies that \citet{zheng2020learning} wrongly claimed an improvement on \citet{zheng2018dags} as they both transition from having no structural knowledge to plausible structural knowledge. Of course, only when we introduce the parametric scale it becomes clear that the difference does not lie in structure, but in their parametric assumptions.

\subsection{The temporal scale} \label{sec:time}

We explicitly differentiate between temporal and static models because the presence of temporal structure may lead to alternative considerations in the causal setting. For example, a typical assumption in the temporal domain is that causes precede effects \citep{granger1969investigating}. While we do not express any opinion on whether or not this makes things harder or easier, we do believe it is important to differentiate to properly compare methods. Furthermore, when using our guideline to search for an appropriate method from a practical point of view, it makes sense to search across static methods for a static problem, and temporal methods for a temporal problem.

\begin{figure}
    \centering
    \begin{tikzpicture}[
        var/.style={circle, inner sep=.2, minimum size=1.5mm}
    ]
        \node[var] (x1) at (0, 0){$X_{t=1}$};
        \node[var] (x2) at ($(x1) + (2, 0)$) {$X_{t=2}$};

        \node[var] (y1) at ($(x1) + (0, -1.5)$) {$Y_{t=1}$};
        \node[var] (y2) at ($(x2) + (0, -1.5)$) {$Y_{t=2}$};

        \node[var] (a1) at ($(x1)!.5!(x2) + (0, -.75)$) {$A_{t=1}$};
        \node[var] (a2) at ($(a1) + (2, 0)$) {$A_{t=2}$};

        \node[var, red] (u1) at ($(a1) + (0, 1.5)$) {$U_{t=1}$};
        \node[var, red] (u2) at ($(a2) + (0, 1.5)$) {$U_{t=2}$};

        \draw[-latex] (x1) -- (x2);
        \draw[-latex] (x1) -- (y1);
        \draw[-latex] (x2) -- (y2);

        \draw[-latex] (x1) -- (a1);
        \draw[-latex] (a1) -- (x2);
        \draw[-latex] (x2) -- (a2);

        \draw[-latex] (x1) -- (u1);
        \draw[-latex] (a1) -- (u1);
        \draw[-latex] (u2) -- (x2);

        \draw[-latex] (u1) -- (x2);
        \draw[-latex, red] (u1) -- (u2);

    \end{tikzpicture}
    \caption{{\bf Confounding over time.} We borrow the above DAG from \citet{bica2020estimating} in two time steps \citet{robins2000marginal}. This figure shows how time complicates matters. Specifically, because of time, the same variable ($X$) has different causal edges associated as a function of $t$. See for example how $X_{t=1}$ causes $U_{t=1}$, but not $U_{t=2}$.}
    \label{fig:time}
    \rule{\linewidth}{.5pt}
\end{figure}
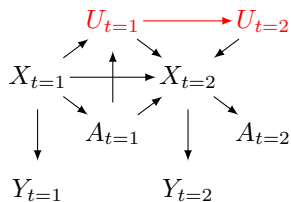

While, in our writing, we consider a model temporal or not to be binary, we wish to emphasise the complexity associated with time.
Especially in settings of causality, we find time to be an incredibly complex and interesting topic.
Consider, for example, \cref{fig:time} which illustrates how a simple time indicator may make a simple four-variable problem a situation with a complex associated DAG.
Interestingly, time can also {\it help} in causality: having $X_{t=1} \to X_{t=2}$ makes much more sense than $X_{t=2} \to X_{t=1}$ because a cause cannot manifest later than an effect.
The latter brings up an interesting thought. If there are causes in a static data setting, they too should manifest {\it before} and effect.
Hence, having static data with an associated causal DAG, always implies there is a temporal character to the data.
As such, we define time here as a setting in which we observe the same variable (such as $X, U, A, Y$ in \cref{fig:time}) more than once.

While \cref{fig:time} is illustrating a discrete-time setting, the above can also be generalised to a continuous time setting.
\citet{peters2022causal} connect the dynamical systems literature to structural causal models.
Modelled as a dynamical system (such as ODEs), we allow the same variable to be observed multiple times while also being dictated by a structural causal model.
Searching for ODEs then connects dynamical systems to a causal structural equation, which translates to an FK model on our parametric scale in \cref{sec:param}.

Now that we have described the three-dimensional scale for characterizing knowledge and assumptions, we will use this scale to characterize, construct, and test CDL model pipelines.

\subsection{Characterizing CDL models using the scale} \label{sec:map}

\begin{figure}[t]
    \centering
    \newcommand\Square[1]{+(-#1,-#1) rectangle +(#1,#1)}
    \begin{tikzpicture}[
        grid/.style={gray, dashed},
        annotation/.style={very thick, red, opacity=.6},
        dot/.style={circle, fill=black, line width=0,  minimum size=2mm, inner sep=0},
        route/.style={very thick, draw opacity=.5},
        scale=.9
    ]
        \node[label={left:Unknown (U)}] (input_knowledge_A) at (0,0) {}; 
        \draw[grid] ($(input_knowledge_A) + (0, .5)$) -- ($(input_knowledge_A) + (4, .5)$);
        \node[label={left:Plausible (P)}] (input_knowledge_U) at ($(input_knowledge_A) + (0, 1)$) {};
        \draw[grid] ($(input_knowledge_U) + (0, .5)$) -- ($(input_knowledge_U) + (4, .5)$);
        \node[label={left:Causal (C)}] (input_knowledge_C) at ($(input_knowledge_U) + (0, 1)$) {};
        
        \node[label=below:(NP)] (input_param_NO) at (.5,-.5) {};
        \node[rotate=-90, anchor=west] at (.5, -1.3) {Non-param.};
        \draw[grid] ($(input_param_NO) + (.5, 0)$) -- ($(input_param_NO) + (.5, 3)$);
        \node[label=below:(NM)] (input_param_1) at ($(input_param_NO) + (1, 0)$) {};
        \node[rotate=-90, anchor=west] at ($(input_param_NO) + (1, -.8)$) {Noise Models};
        \draw[grid] ($(input_param_1) + (.5, 0)$) -- ($(input_param_1) + (.5, 3)$);
        \node[label=below:(Pa.)] (input_param_2) at ($(input_param_1) + (1, 0)$) {};
        \node[rotate=-90, anchor=west] at ($(input_param_1) + (1, -.8)$) {Param.};
        \draw[grid] ($(input_param_2) + (.5, 0)$) -- ($(input_param_2) + (.5, 3)$);
        \node[label={below:{(FK)}}] (input_param_M) at ($(input_param_2) + (1, 0)$) {};
        \node[rotate=-90, anchor=west] at ($(input_param_2) + (1, -.8)$) {Fully known};

        \draw[very thick] ($(input_knowledge_A) - (0, .5)$) -- ($(input_knowledge_C) + (0, .5)$) -- ($(input_knowledge_C) + (4, .5)$) -- ($(input_param_M) + (.5, 0)$) -- cycle;
        
        \node (input_title) at ($(input_param_NO)!.5!(input_param_M) + (0, 3.5)$) {\bf A Priori Knowledge};

        
        \node[inner sep=0] (time_start) at ($(input_param_M) + (1.5, 3.5)$) {};
        \node[inner sep=0] (time_end) at ($(input_param_M) + (1.5, 1.5)$) {};
        \node[inner sep=0] (static_start) at ($(input_param_M) + (1.5, .5)$) {};
        \node[inner sep=0] (static_end) at ($(input_param_M) + (1.5, -1.5)$) {};
        
        \draw[very thick, blue] (time_start) -- (time_end);
        \draw[very thick, blue] (static_start) -- (static_end);
        
        \node at ($(time_start) + (0, .3)$) {\bf \textcolor{blue}{Time}};
        \node at ($(static_start) + (0, .3)$) {\bf \textcolor{blue}{Static}};

        \node[label={left:U}] (repr_knowledge_A) at ($(input_knowledge_A) + (7, 0)$) {}; 
        \draw[grid] ($(repr_knowledge_A) + (0, .5)$) -- ($(repr_knowledge_A) + (4, .5)$);
        \node[label={left:P}] (repr_knowledge_U) at ($(repr_knowledge_A) + (0, 1)$) {};
        \draw[grid] ($(repr_knowledge_U) + (0, .5)$) -- ($(repr_knowledge_U) + (4, .5)$);
        \node[label={left:C}] (repr_knowledge_C) at ($(repr_knowledge_U) + (0, 1)$) {};
        
        \node[label={below:NP}] (repr_param_NO) at ($(repr_knowledge_A) + (.5, -.5)$) {};
        \draw[grid] ($(repr_param_NO) + (.5, 0)$) -- ($(repr_param_NO) + (.5, 3)$);
        \node[label={below:NM}] (repr_param_1) at ($(repr_param_NO) + (1, 0)$) {};
        \draw[grid] ($(repr_param_1) + (.5, 0)$) -- ($(repr_param_1) + (.5, 3)$);
        \node[label={below:Pa.}] (repr_param_2) at ($(repr_param_1) + (1, 0)$) {};
        \draw[grid] ($(repr_param_2) + (.5, 0)$) -- ($(repr_param_2) + (.5, 3)$);
        \node[label={below:FK}] (repr_param_M) at ($(repr_param_2) + (1, 0)$) {};

        \draw[very thick] ($(repr_knowledge_A) - (0, .5)$) -- ($(repr_knowledge_C) + (0, .5)$) -- ($(repr_knowledge_C) + (4, .5)$) -- ($(repr_param_M) + (.5, 0)$) -- cycle;
        
        \node (repr_title) at ($(repr_param_NO)!.5!(repr_param_M) + (0, 3.5)$) {\bf A Posteriori Knowledge};

        
        \draw[annotation] ($(input_title) - (1.55, .2)$) -- ($(input_title) + (1.55, -.2)$);
        \draw[->, annotation] ($(input_title) - (0, .2)$) to[out=270, in=0] ($(input_title) - (3, 0)$) node[annotation, text width=3cm, align=right, anchor=east] {properties of the data {\it given} to a model};

        \draw[annotation] ($(repr_title) - (1.9, .2)$) -- ($(repr_title) + (1.9, -.2)$);
        \draw[->, annotation] ($(repr_title) - (0, .2)$) to[out=270, in=90] ($(repr_title) + (3.7, -2)$) node[annotation, text width=3cm, align=center, anchor=north] {properties of the model's representation or output};
        
        \draw[latex-latex, annotation] ($(repr_knowledge_C) + (-.1, .5)$) -- ($(repr_knowledge_C) + (-.7, .5)$) -- ($(repr_knowledge_A) + (-.7, -.5)$) -- ($(repr_knowledge_A) + (-.1, -.5)$);
        \draw[->, annotation] ($(repr_knowledge_C)!.6!(repr_knowledge_A) + (-.7, 0)$) to[out=180, in=90] ($(repr_param_NO) + (-1, -2)$) node[annotation, text width=4cm, align=center, anchor=north] {structural scale (\cref{fig:scale})};
        
        \draw[latex-latex, annotation] ($(repr_param_NO) - (.5, .1)$) -- ($(repr_param_NO) - (.5, .7)$) -- ($(repr_param_M) + (.5, -.7)$) -- ($(repr_param_M) + (.5, -.1)$);
        \draw[->, annotation] ($(repr_param_NO)!.5!(repr_param_M) + (0,-.7)$) to[out=270, in=180] ($(repr_param_M) + (.5, -1.7)$) node[annotation, text width=3cm, align=left, anchor=west] {parametric scale (\cref{fig:param})};

        \draw[gray, route] ($(input_knowledge_A) + (.5, 0) + 1*(1, 0)$) node[dot, gray] {} -- node[midway, above, rotate=30, gray, opacity=1] {Example model} ($(time_start) - (0, 1)$) node[dot, gray] {} -- ($(repr_knowledge_C) + (.5, 0) + 1*(1, 0)$) node[dot, gray] {};
            
    \end{tikzpicture}
    \caption{{\bf Combining axes.} We can combine our axes to evaluate a model's input and output (or representation). As an example, we show a fictitious method which assumes no structure but {\it does} make an additive noise assumption; operates in the temporal domain; and provides a truly causal representation under the same additive noise assumption. Doing so allows easy examination of a method on many properties at once.}
    \label{fig:map:annotated}
    \rule{\textwidth}{.5pt}

\end{figure}
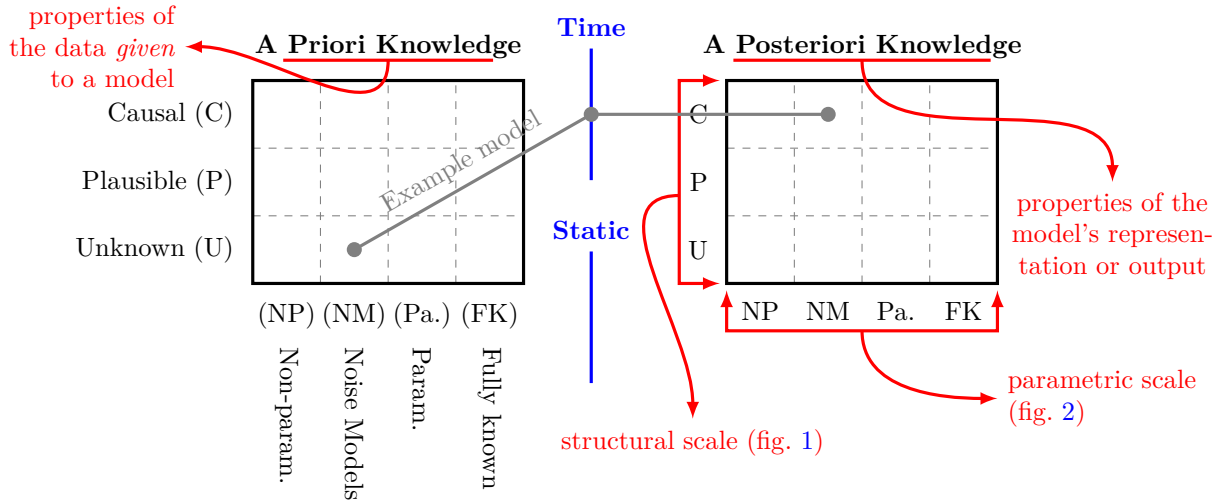

Using our scales in \cref{fig:scale,fig:param}, we provide an organisational structure which allows comparing and documenting CDL methods. 
Each CDL method can be characterized by answering the following three questions:
\begin{itemize}
    \item What is the a priori knowledge (or assumptions) according to the structural and parametric scale?
    \item Is the model static or temporal?
    \item What is the a posteriori knowledge after applying the model to data?
\end{itemize}
This framework is illustrated in Figure \ref{fig:map:annotated}.

{\bf A priori knowledge.} The a priori knowledge field (leftmost grid in \cref{fig:map:annotated}) reflects the structural and parametric knowledge that the practitioner possesses or is willing to assume before applying the CDL method to data. 
Here, the structural scale measures how much we already know about {\it which} variables are dependent; whereas the parametric scale measures {\it how} these variables are dependent. 

{\bf Time.} This is a binary indicator of whether the CDL method (and the dataset) is static or temporal. 

{\bf A posteriori knowledge.} The a posteriori knowledge field reflects the structural and parametric knowledge that one would obtain after applying the CDL method to the data. 
It is evaluated using the same scales as the a priori knowledge, which helps to document how much additional knowledge the CDL modelling process adds.

With our framework, we can fingerprint each CDL method based on its location on the map. As we will show in the next sections, this framework not only allows us to categorize model classes but also to construct and test new CDL model pipelines. 

\section{Constructing and testing model pipelines} \label{sec:transitions}

\subsection{Transition in the knowledge}

We recognize that a CDL method's a priori and a posteriori knowledge may be different because the method may discover additional knowledge from the data. We use the term Transition to describe this situation. 
\begin{definition}[Transition]\label{def:transition}
When a method's a priori and a posteriori knowledge is different, either structural (\cref{sec:scale}) or parametric (\cref{sec:param}), we say that a model invokes a transition in the knowledge. 
\end{definition}
A transition may occur either in the structural scale or parametric scale (or both).

\textbf{Structural Transition}. Causal learning is a prime example of a structural transition, where the input structure (typically unknown) is at a lower level than the output (a causal graph). This applies to various causal structural learning methods that learn (or discover) a causal graph from data.

\textbf{Parametric Transition}. Transition on the parametric scale often involves testing functional form (e.g. linearity) or the distribution of the noise terms (e.g. Gaussian noise). Symbolic regression methods, which try to discover closed-form equations from data, would also invoke a transition in the parametric scale. 

In our \cref{sec:nav} we categorise existing models in supervised learning, forecasting, bandit algorithms, reinforcement learning, and generative modelling. 
In none of these methods we find transitions that end up in a fully known (FK) structural equation model. 
Clearly this type of transition is an interesting avenue to explore further in research.

In our treatment effects example in \cref{sec:TEs} we perform a similar exercise and find one model \citep{holt2024ode} (to our knowledge) that {\it does} transition in the parametric scale with a FK equation model as posteriori knowledge.

\textbf{No Transition}. In general, Causal reasoning, such as counterfactual prediction and treatment effect estimation, constitutes {\it no} transition as the algorithms aim to estimate the causal effect rather than discovering structural or parametric properties. Examples of this include bandit algorithms such as \citet{lattimore2016causal}, which we discuss further using our map of CDL in \cref{sec:nav}.

\begin{figure}[t]
    \hspace*{-.8cm}
    \newcommand\Square[1]{+(-#1,-#1) rectangle +(#1,#1)}
    \begin{tikzpicture}[
        grid/.style={gray, dashed},
        annotation/.style={very thick, red, opacity=.6},
        dot/.style={circle, fill=black, line width=0,  minimum size=2mm, inner sep=0},
        route/.style={very thick, draw opacity=.5},
        scale=.8
    ]
        \node[label={left:\small Unknown (U)}] (input_knowledge_A) at (0,0) {}; 
        \draw[grid] ($(input_knowledge_A) + (0, .5)$) -- ($(input_knowledge_A) + (4, .5)$);
        \node[label={left:\small Plausible (P)}] (input_knowledge_U) at ($(input_knowledge_A) + (0, 1)$) {};
        \draw[grid] ($(input_knowledge_U) + (0, .5)$) -- ($(input_knowledge_U) + (4, .5)$);
        \node[label={left:\small Causal (C)}] (input_knowledge_C) at ($(input_knowledge_U) + (0, 1)$) {};
        
        \node[label=below:\small (NP)] (input_param_NO) at (.5,-.5) {};
        \node[rotate=-90, anchor=west] at (.5, -1.3) {\small Non-param.};
        \draw[grid] ($(input_param_NO) + (.5, 0)$) -- ($(input_param_NO) + (.5, 3)$);
        \node[label=below:\small (NM)] (input_param_1) at ($(input_param_NO) + (1, 0)$) {};
        \node[rotate=-90, anchor=west] at ($(input_param_NO) + (1, -.8)$) {\small Noise Models};
        \draw[grid] ($(input_param_1) + (.5, 0)$) -- ($(input_param_1) + (.5, 3)$);
        \node[label=below:\small (Pa.)] (input_param_2) at ($(input_param_1) + (1, 0)$) {};
        \node[rotate=-90, anchor=west] at ($(input_param_1) + (1, -.8)$) {\small Param.};
        \draw[grid] ($(input_param_2) + (.5, 0)$) -- ($(input_param_2) + (.5, 3)$);
        \node[label={below:{\small (FK)}}] (input_param_M) at ($(input_param_2) + (1, 0)$) {};
        \node[rotate=-90, anchor=west] at ($(input_param_2) + (1, -.8)$) {\small Fully known};

        \draw[very thick] ($(input_knowledge_A) - (0, .5)$) -- ($(input_knowledge_C) + (0, .5)$) -- ($(input_knowledge_C) + (4, .5)$) -- ($(input_param_M) + (.5, 0)$) -- cycle;
        
        \node (input_title) at ($(input_param_NO)!.5!(input_param_M) + (0, 3.7)$) {\bf 1\textsuperscript{st} Cascade};

        \node[label={left:\small U}] (repr_knowledge_A) at ($(input_knowledge_A) + (6, 0)$) {}; 
        \draw[grid] ($(repr_knowledge_A) + (0, .5)$) -- ($(repr_knowledge_A) + (4, .5)$);
        \node[label={left:\small P}] (repr_knowledge_U) at ($(repr_knowledge_A) + (0, 1)$) {};
        \draw[grid] ($(repr_knowledge_U) + (0, .5)$) -- ($(repr_knowledge_U) + (4, .5)$);
        \node[label={left:\small C}] (repr_knowledge_C) at ($(repr_knowledge_U) + (0, 1)$) {};
        
        \node[label={below:\small NP}] (repr_param_NO) at ($(repr_knowledge_A) + (.5, -.5)$) {};
        \draw[grid] ($(repr_param_NO) + (.5, 0)$) -- ($(repr_param_NO) + (.5, 3)$);
        \node[label={below:\small NM}] (repr_param_1) at ($(repr_param_NO) + (1, 0)$) {};
        \draw[grid] ($(repr_param_1) + (.5, 0)$) -- ($(repr_param_1) + (.5, 3)$);
        \node[label={below:\small Pa.}] (repr_param_2) at ($(repr_param_1) + (1, 0)$) {};
        \draw[grid] ($(repr_param_2) + (.5, 0)$) -- ($(repr_param_2) + (.5, 3)$);
        \node[label={below:\small FK}] (repr_param_M) at ($(repr_param_2) + (1, 0)$) {};

        \draw[very thick] ($(repr_knowledge_A) - (0, .5)$) -- ($(repr_knowledge_C) + (0, .5)$) -- ($(repr_knowledge_C) + (4, .5)$) -- ($(repr_param_M) + (.5, 0)$) -- cycle;
        
        \node (repr_title) at ($(repr_param_NO)!.5!(repr_param_M) + (0, 3.7)$) {\bf 2\textsuperscript{nd} Cascade};

        \node[label={left:\small U}] (casc_knowledge_A) at ($(repr_knowledge_A) + (6, 0)$) {}; 
        \draw[grid] ($(casc_knowledge_A) + (0, .5)$) -- ($(casc_knowledge_A) + (4, .5)$);
        \node[label={left:\small P}] (casc_knowledge_U) at ($(casc_knowledge_A) + (0, 1)$) {};
        \draw[grid] ($(casc_knowledge_U) + (0, .5)$) -- ($(casc_knowledge_U) + (4, .5)$);
        \node[label={left:\small C}] (casc_knowledge_C) at ($(casc_knowledge_U) + (0, 1)$) {};
        
        \node[label={below:\small NP}] (casc_param_NO) at ($(casc_knowledge_A) + (.5, -.5)$) {};
        \draw[grid] ($(casc_param_NO) + (.5, 0)$) -- ($(casc_param_NO) + (.5, 3)$);
        \node[label={below:\small NM}] (casc_param_1) at ($(casc_param_NO) + (1, 0)$) {};
        \draw[grid] ($(casc_param_1) + (.5, 0)$) -- ($(casc_param_1) + (.5, 3)$);
        \node[label={below:\small Pa.}] (casc_param_2) at ($(casc_param_1) + (1, 0)$) {};
        \draw[grid] ($(casc_param_2) + (.5, 0)$) -- ($(casc_param_2) + (.5, 3)$);
        \node[label={below:\small FK}] (casc_param_M) at ($(casc_param_2) + (1, 0)$) {};

        \draw[very thick] ($(casc_knowledge_A) - (0, .5)$) -- ($(casc_knowledge_C) + (0, .5)$) -- ($(casc_knowledge_C) + (4, .5)$) -- ($(casc_param_M) + (.5, 0)$) -- cycle;
        
        \node (repr_title) at ($(casc_param_NO)!.5!(casc_param_M) + (0, 3.7)$) {\bf 3\textsuperscript{rd} Cascade};

        \node[dot, blue] (start_goal_model) at ($(input_knowledge_A) + (.5, 0) + 1*(1, 0) + (.1, -.25)$) {};
        \node[dot, blue] (end_goal_model) at ($(casc_knowledge_C) + (.5, 0) + 1*(1, 0) + (-.1, -.25)$) {};
        
        \node[dot, red] (start_trans_model) at ($(start_goal_model) + (0, .5)$) {};
        \node[dot, red] (end_trans_model) at ($(repr_knowledge_C) + (.5, 0) + 1*(1, 0) + (0, -.25)$) {};
        
        \draw[densely dotted, thick, ->] (start_goal_model) to[out=180, in=180, looseness=2] (start_trans_model);
        
        \node[dot, DarkGreen] (start_original_model) at ($(end_trans_model) - (1, 0) + (.25, .45)$) {};
        \node[dot, DarkGreen] (end_original_model) at ($(end_goal_model) + (0, .45)$) {};

        \draw[red, route] (start_trans_model) -- node[pos=.45, above, rotate=15, red, opacity=1, fill=white, inner sep=.5] {\bf Causal learning \citep{peters2014jmlr}}  (end_trans_model) node[dot, red] {};
        \draw[DarkGreen, route] (start_original_model) -- node[pos=.5, above, rotate=0, DarkGreen, opacity=1, fill=white, inner sep=.5] {\bf Causal debiasing \citep{van2021decaf}}  (end_original_model) node[dot, DarkGreen] {};
        \draw[blue, route] (start_goal_model) -- node[pos=.55, below, rotate=9.5, blue, opacity=1, fill=white, inner sep=.5] {\bf Generate fair data} (end_goal_model) node[dot, blue] {};

        \draw[densely dotted, thick, ->] (end_trans_model) -- (start_original_model);
        \draw[densely dotted, thick, ->] (end_original_model) to[out=0, in=0, looseness=2] (end_goal_model);

    \end{tikzpicture}
    \caption{{\bf Building pipelines.} As the input and representation in our map (\cref{fig:map:annotated}) is defined on the same domain-- the structural scale $\times$ the parametric scale --we can use the representation of a model as input for a subsequent model. Documenting these model combinations (i.e. pipelines) is easy with our scales as we can simply cascade multiple diagrams after one another as we have done above. Doing so shows us that, to truly generate fair data, one has to be willing to make either strong assumptions (such as having access to causal DAGs, or the necessary assumptions to discover them), or use alternative models in their pipeline that make testable assumptions.} \label{fig:cascading}
    \rule{\linewidth}{.5pt}
\end{figure}
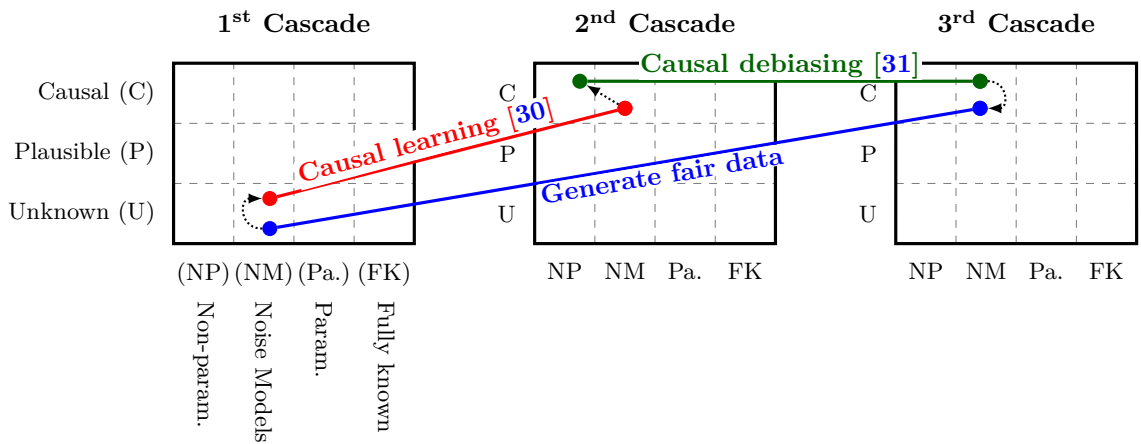

\subsection{Building CDL model pipelines}

Very often a standalone CDL model cannot solve the practical problem because there is a gap between a priori knowledge of the practitioner and what is required by the model. With CDL, we can identify the right type of transition and construct model pipelines to bridge the knowledge gap. We consider the example below (visualized in Fig \ref{fig:cascading}). 

\citet{van2021decaf} proposed a GAN which generates fair synthetic data by exploiting a known causal graph. While useful, it is hard to apply the algorithm in practice as a fully known causal graph is seldom available in many application domains.

From the map, it is clear that the practitioner needs to incorporate a transition model as \citet{van2021decaf} (illustrated in green) cannot function with the practitioner's insufficient knowledge of the causal structure. As such, we first learn a causal graph, using for example \citet{peters2014jmlr}.
With the two-stage model pipeline, we can match the a priori knowledge of the practitioner.

Naturally, our illustration is quite straightforward and one could likely arrive at the same solution without our map. However, often real-world scenarios can be much more complicated and require more than two cascading models to solve a particular problem \citep{berrevoets21a,kunft2019intermediate}. Imagine problems such as data imputation, transformation, or life-long learning. All of these intermediary steps have to take into account the assumptions made on each model present in the pipeline. Building on a set of non-cascading assumptions can have detrimental consequences as there may be unwanted bias creeping into your solution.

For example, in the above scenario, the observational data used to learn a causal graph may have missing values. Depending on which imputation strategy we use, we may introduce additional parametric assumptions into our pipeline. Furthermore, depending on which mechanism governs the missingness patterns, we may have to make additional structural assumptions before imputation can even begin \citep{berrevoets2022impute,mohan2021graphical,mayer2020doubly}. Making such structural assumptions, {\it before} learning a graph using \citet{peters2014jmlr} would make our solution for the practitioner's problem invalid once again.

\subsection{Non-matching assumptions}

In our example in \cref{fig:cascading}, we have not addressed the fact that the used structure learner yields a graph on a different parametric level than what is assumed by the reasoning task. Namely, RESIT leans a causal structure assuming noise model structural equations, while \citet{lattimore2016causal} assumes a non-parametric causal graph. 

Luckily, because of the composition of both the structural and parametric scales, each stricter assumption is subsumed by a lower-level assumption. As such, from left to right, we can always relax the assumptions (such as allowing non-parametric reasoning based on a noise-model causal structure), but not the other way around. If, for example, our structure learner yielded a plausible but non-parametric causal structure, we are not guaranteed optimal regret from \citet{lattimore2016}. On our map, that would become clear as we would move from a less strict assumption (plausible causality) to a strict assumption (full causality).

\section{Treatment effects: An illustrative example using CDL} \label{sec:TEs}

We have seen that the CDL framework allows practitioners to verify the suitability of a model and build model pipelines. Let us discuss how we can arrive at such a conclusion using a practical example of treatment effects. In particular, we will focus our CDL efforts on solving the common case of inferring a treatment effect from the infant health and development program (IHDP) dataset \citep{hill2011bayesian,curth2021really}.

Applying treatment effects in practice, the CDL way amounts to the following steps:
(1) {\it A priori knowledge:} Identify the structural and parametric properties of the data.
(2) {\it Temporal or static:} Is the problem we wish to solve one that requires inference over time, or not?
(3) {\it Posteriori knowledge:} What type of structural and parametric should our resulting model have?

{\bf A brief overview of CATE.} Before applying our framework as laid out above, let us first briefly recapitulate the general treatment effects setting, in particular {\it conditional average treatment effect} (CATE) inference, which focuses on the following object:
\begin{equation} \label{eq:cate}
    \tau(X) \coloneqq \mathbf{E}[Y_1 - Y_0 | X],
\end{equation}
where $Y_A$ (with $A \in \{0, 1\}$) is the {\it potential outcome} given a treatment $A$ (which in \cref{eq:cate} is binary).
\Cref{eq:cate} is a {\it conditional} ATE due to $X$. If $X$ is dropped from \cref{eq:cate} we are left with an ATE which covers the complete population $\mathcal{X}$ with $X\in\mathcal{X}$.
Interestingly, our objective, $\tau(X)$, is impossible to observe! We can only observe parts of this: $\mathbb{E}[Y_A | X]$, again with $A \in \{0,1\}$.
The tricky thing about only observing one term in the expectation in \cref{eq:cate}, is that the per-term expectations are covering possible different populations.
In particular, we have:
\begin{equation*}
    p(X | A=1) \neq p(X | A=0),
\end{equation*}
which does not allow to split \cref{eq:cate} into a difference between $\mathbb{E}[Y_1 | X]$ and $\mathbb{E}[Y_0 | X]$-- that right is reserved only for a randomised trial (with $p(X | A=1) = p(X | A=0)$ for all $X \in \mathcal{X}$) \citep{pearl2018book}.

There are several strategies to infer $\tau(X)$ regardless, either directly or in parts by inferring each potential outcome separately.
For an overview of different inference tactics, we refer to \citet{curth2021nonparametric,kunzel2019metalearners}.

\subsection{Treatment effects: the CDL description}
Let us use our CDL perspective to describe what is required to solve the treatment effects problem given the IHDP dataset \citep{hill2011bayesian}.

Treatment effects papers make three starting assumptions: Overlap (each individual in the data has a non-zero probability of receiving treatment), consistency (the observed outcome corresponds with a potential outcome), and ignorability (all variables that confound treatment and the potential outcomes are observed).
The last assumption is often expressed as: $A \independent \{Y_a : a \in \mathcal{A}\} | X$, where $X$ are the individual's covariates, $A \in \mathcal{A}$ is the treatment, and $\{Y_a : a \in \mathcal{A}\}$ is the set of potential outcomes \cite{neyman1923applications,rubin80comment}.
The complication from the above setup is that one can only observe {\it one} potential outcome \citep{holland1986statistics}.
Hence, the above conditional independence statement can never be observed (as it concerns the {\it complete} set of potential outcomes).

\begin{element}{A Priori Knowledge}

    {\bf The structural scale.} Translating this conditional independence statement to a causal structure, would look like $A \leftarrow X \rightarrow \{Y_a : a \in \mathcal{A}\}$ \citep{shalit2017estimating}.
    This is a structural causal assumption \citep{richardson2013single}. However, because of the special nature of the potential outcomes framework (where only one of the potential outcomes can be observed), one cannot validate this causal statement, even with an elaborate experimental design \citep{pearl2018book}.

    {\bf The parametric scale.} From IHDP's description we know that the potential outcome surfaces follow the following functions:
    \begin{equation*}
        \mu_0 = \exp((X + M)\beta) \quad \text{and} \quad \mu_1 = X\beta + \omega,
    \end{equation*}
    with $\mu_a$ the expectation of $Y_a$ given $X$.
    Given the above outcome surfaces, it would make sense to use a linear estimator for $\mu_1$, and an exponential estimator for $\mu_0$ (or even a linear one for $\ln \mu_0$).
    Interestingly, most work that currently uses IHDP as a benchmark dataset mostly focuses on neural networks, {\it which is situated two places left on the structural scale!}

\end{element}

\begin{figure}[t]
    \centering
    \hspace*{-1.2cm}  
    \begin{tikzpicture}[
        dot/.style={circle, fill=black, line width=0,  minimum size=2mm, inner sep=0},
        grid/.style={gray, dashed},
        route/.style={very thick, draw opacity=.5},
        bandit/.style={dashed},
        scale=.9
    ]
        
        \node[label={left:U}] (input_knowledge_A) at (0,0) {}; 
        \draw[grid] ($(input_knowledge_A) + (0, .5)$) -- ($(input_knowledge_A) + (4, .5)$);
        \node[label={left:P}] (input_knowledge_U) at ($(input_knowledge_A) + (0, 1)$) {};
        \draw[grid] ($(input_knowledge_U) + (0, .5)$) -- ($(input_knowledge_U) + (4, .5)$);
        \node[label={left:C}] (input_knowledge_C) at ($(input_knowledge_U) + (0, 1)$) {};
        
        \node[label={below:NP}] (input_param_NO) at (.5,-.5) {};
        \draw[grid] ($(input_param_NO) + (.5, 0)$) -- ($(input_param_NO) + (.5, 3)$);
        \node[label={below:NM}] (input_param_1) at ($(input_param_NO) + (1, 0)$) {};
        \draw[grid] ($(input_param_1) + (.5, 0)$) -- ($(input_param_1) + (.5, 3)$);
        \node[label={below:Pa.}] (input_param_2) at ($(input_param_1) + (1, 0)$) {};
        \draw[grid] ($(input_param_2) + (.5, 0)$) -- ($(input_param_2) + (.5, 3)$);
        \node[label={below:FK}] (input_param_M) at ($(input_param_2) + (1, 0)$) {};

        \draw[very thick] ($(input_knowledge_A) - (0, .5)$) -- ($(input_knowledge_C) + (0, .5)$) -- ($(input_knowledge_C) + (4, .5)$) -- ($(input_param_M) + (.5, 0)$) -- cycle;
        
        \node at ($(input_param_NO)!.5!(input_param_M) + (0, 3.5)$) {\bf A Priori Knowledge};

        
        \node[inner sep=0] (time_start) at ($(input_param_M) + (2, 4.5)$) {};
        \node[inner sep=0] (time_end) at ($(input_param_M) + (2, 2.5)$) {};
        \node[inner sep=0] (static_start) at ($(input_param_M) + (2, 1.5)$) {};
        \node[inner sep=0] (static_end) at ($(input_param_M) + (2, -.5)$) {};
        
        \draw[very thick, blue] (time_start) -- (time_end);
        \draw[very thick, blue] (static_start) -- (static_end);
        
        \node at ($(time_start) + (0, .3)$) {\bf \textcolor{blue}{Time}};
        \node at ($(static_start) + (0, .3)$) {\bf \textcolor{blue}{Static}};

        
        \node (repr_knowledge_A) at ($(input_knowledge_A) + (7, 0)$) {}; 
        \draw[grid] ($(repr_knowledge_A) + (0, .5)$) -- ($(repr_knowledge_A) + (4, .5)$);
        \node (repr_knowledge_U) at ($(repr_knowledge_A) + (0, 1)$) {};
        \draw[grid] ($(repr_knowledge_U) + (0, .5)$) -- ($(repr_knowledge_U) + (4, .5)$);
        \node (repr_knowledge_C) at ($(repr_knowledge_U) + (0, 1)$) {};
        
        \node[label={below:NP}] (repr_param_NO) at ($(repr_knowledge_A) + (.5, -.5)$) {};
        \draw[grid] ($(repr_param_NO) + (.5, 0)$) -- ($(repr_param_NO) + (.5, 3)$);
        \node[label={below:NM}] (repr_param_1) at ($(repr_param_NO) + (1, 0)$) {};
        \draw[grid] ($(repr_param_1) + (.5, 0)$) -- ($(repr_param_1) + (.5, 3)$);
        \node[label={below:Pa.}] (repr_param_2) at ($(repr_param_1) + (1, 0)$) {};
        \draw[grid] ($(repr_param_2) + (.5, 0)$) -- ($(repr_param_2) + (.5, 3)$);
        \node[label={below:FK}] (repr_param_M) at ($(repr_param_2) + (1, 0)$) {};

        \draw[very thick] ($(repr_knowledge_A) - (0, .5)$) -- ($(repr_knowledge_C) + (0, .5)$) -- ($(repr_knowledge_C) + (4, .5)$) -- ($(repr_param_M) + (.5, 0)$) -- cycle;
        
        \node at ($(repr_param_NO)!.5!(repr_param_M) + (0, 3.5)$) {\bf A Posteriori Knowledge};


        \draw[DarkTurquoise, route, bandit] ($(input_knowledge_C) + (.5, .25) + 0*(1, 0)$)  node[dot, DarkTurquoise] {} to[out=45, in=180] ($(time_start)!.25!(time_end)$) node[dot, DarkTurquoise] {} to[out=0, in=115 ] ($(repr_knowledge_C) + (.5, .25) + 0*(1, 0)$) node[dot, DarkTurquoise] {};
        \node[label={[text=DarkTurquoise]right:\citep{bica2020estimating,berrevoets2021disentangled,lim2018forecasting,vanderschueren23a}}] at ($(repr_knowledge_C) + (4.5, .25)$) {\textcolor{DarkTurquoise}{\dott}};

        \draw[orange, route, bandit] ($(input_knowledge_C) + (.5, .25) + 1*(1, 0)$) node[dot, orange] {} to[out=25, in=180] ($(time_start)!.5!(time_end)$) node[dot, orange] {} to[out=0, in=155] ($(repr_knowledge_C) + (.5, .25) + 1*(1, 0)$) node[dot, orange] {};
        \node[label={[text=orange]right:\citep{seedat2022continuous}}] at ($(repr_knowledge_C) + (7, .25)$) {\textcolor{orange}{\dott}};

        \draw[purple, route, bandit] ($(input_knowledge_C) + (.5, .25) + 2*(1, 0)$) node[dot, purple] {} to[out=5, in=180] ($(time_start)!.75!(time_end)$) node[dot, purple] {} to[out=0, in=175] ($(repr_knowledge_C) + (.5, .25) + 2*(1, 0)$) node[dot, purple] {};
        \node[label={[text=purple]right:\citep{robins2000marginal}}] at ($(repr_knowledge_C) + (8.2, .25)$) {\textcolor{purple}{\dott}};


        \draw[Goldenrod, route] ($(input_knowledge_C) + (.5, -.25) + 0*(1, 0)$)  node[dot, Goldenrod] {} to [out=-45, in=180] ($(static_start)!.75!(static_end)$) node[dot, Goldenrod] {} to[out=0, in=225]  ($(repr_knowledge_C) + (.5, -.25) + 0*(1, 0)$) node[dot, Goldenrod] {};
        \node[label={[text=Goldenrod]right:\citep{berrevoets2020organite,curth2021nonparametric,yoon2018ganite,vanderschueren2023noflite,bica2020scigan}}] at ($(repr_knowledge_C) + (4.5, -.25)$) {\textcolor{Goldenrod}{\dott}};
        
        \draw[blue, route] ($(input_knowledge_C) + (.5, -.25) + 1*(1, 0)$)  node[dot, blue] {} to [out=-45, in=180] ($(static_start)!.5!(static_end)$) node[dot, blue] {} to[out=0, in=225]  ($(repr_knowledge_C) + (.5, -.25) + 1*(1, 0)$) node[dot, blue] {};
        \node[label={[text=blue]right:\citep{berrevoets2021learning,alaa2017bayesian}}] at ($(repr_knowledge_C) + (7, -.25)$) {\textcolor{blue}{\dott}};

        \draw[Violet, route] ($(input_knowledge_C) + (.5, -.25) + 2*(1, 0)$)  node[dot, Violet] {} to [out=-45, in=180] ($(static_start)!.25!(static_end)$) node[dot, Violet] {} to[out=0, in=225]  ($(repr_knowledge_C) + (.5, -.25) + 2*(1, 0)$) node[dot, Violet] {};
        \node[label={[text=Violet]right:\citep{berrevoets2022impute}}] at ($(repr_knowledge_C) + (9, -.25)$) {\textcolor{Violet}{\dott}};

        \draw[DarkGreen, route, bandit] ($(input_knowledge_C) + (.75, 0) + 2*(1, 0)$)  node[dot, DarkGreen] {} to [out=70, in=180] ($(time_start)!.05!(time_end)$) node[dot, DarkGreen] {} to[out=0, in=-225]  ($(repr_knowledge_C) + (.5, 0) + 3*(1, 0)$) node[dot, DarkGreen] {};
        \node[label={[text=DarkGreen]right:\citep{holt2024ode}}] at ($(repr_knowledge_C) + (9.5, .25)$) {\textcolor{DarkGreen}{\dott}};

   \end{tikzpicture}
   
   \caption{{\bf Treatment effects and CDL.} We document some of our own, and other well-known, treatment effects papers in the diagram presented in \cref{sec:transitions}. The basic takeaways from doing this are: (1) Learning treatment effects models requires making causal assumptions about the data-generating process, (2) The focus of the treatment effects community has been relaxing parametric assumptions, (3) learning fully known functions (FK) has not been a focus in this community. In \cref{sec:cartography} we document many more papers across many more fields.
   Here too: Non-parametric (NP), Noise models (NM), Parametric (Pa.), and Fully known (FK); and Unknown (U), Plausible (P), and Causal (C).}
   \label{fig:TEs}
   \rule{\textwidth}{.5pt}
\end{figure}
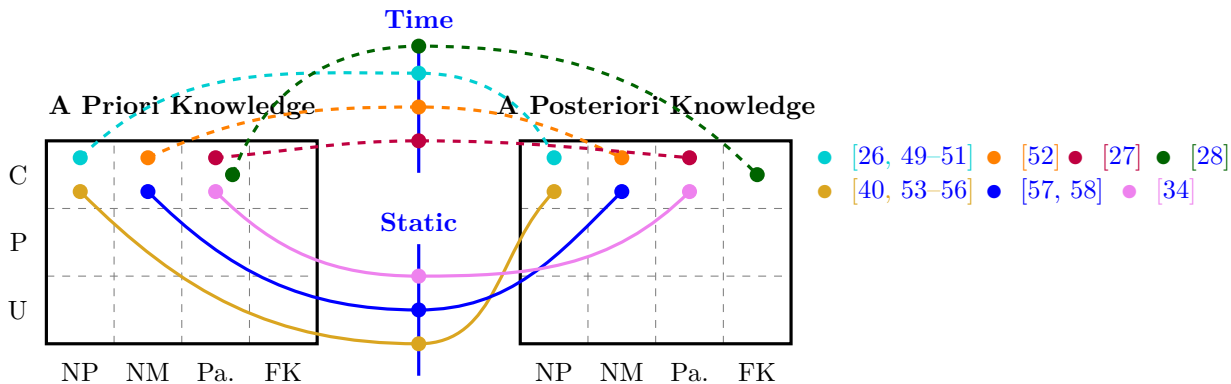

Given the above description, we argue that treatment effects papers which assume the standard set of the {\it stable unit treatment value} assumptions (SUTVA) from above \citep{rubin1990commentneyman}, make causal structural assumptions as is illustrated in \cref{fig:TEs}.

Recall from above that temporal datasets are those that observe an instance of a variable multiple times. 
That means that datasets which simply include a timestamp, are not classified as a temporal dataset.

\begin{element}{Time}
    Hence, from this definition, we know that IHDP is not a temporal dataset, and thus does not require a model that explicitly models for time. In our \cref{fig:TEs}, that means we should not consider the models that are illustrated by a dashed line.
\end{element}

We find that CATE models (and TE models in general \citep{imbens2015causal}) are not concerned with transitioning across structural models.
That is, a CATE model does typically not learn additional causal structure on top of the structure it assumes a priori.

\begin{element}{Posteriori Knowledge}
    A CATE model's structure present in its representation remains the same as those assumed at the input, i.e. the structural assumptions made a priori.
    As such, CATE models typically do not {\it add} causal knowledge on top of what is assumed at the start.
\end{element}

From a practical perspective, the reason for this is that a CATE-model's goal is not to determine causality, but instead to infer a causal {\it effect} \citep{imbens2020potential}.

We hope that the above provides an example of what we advocate any CDL paper to include (a clear description of the problem and solution spaces discussed in the paper using our CDL axes), in some form or another.
We have performed this exercise for a few well-known treatment effects models in \cref{fig:TEs}. 
Interestingly, only in temporal models did we find a treatment effects models that makes a transition on the parametric scale, ending up in a fully known (FK) parametric description. 
That is because, in \citet{holt2024ode}, the objective is to explicitly learn underlying equations which are later used for treatment effects inference.

For more practical ML examples, we refer to our \cref{sec:RW} which lists a few where we observe CDL to already have made some impact.

\section{Conclusion and guidelines for CDL papers} \label{sec:discussion}

In conclusion, our framework for causal deep learning (CDL) represents a major step forward in the field of machine learning research. 
By incorporating a structural dimension to measure absent, present, or partially known structural causal knowledge in methods, a parametric dimension that maps the strengths of deep learning methods to causality, and a temporal dimension that accounts for the impact of time on a causal system, we have provided a comprehensive guide for using CDL to solve real-world problems.

Our framework extends the levels of structural knowledge in causality by introducing a new level, "plausible causal knowledge", that accommodates many existing papers and provides a clear pathway to powerful tools and methods. 
With our clear definition, we can now identify real problems and select the method that best fits the setting of interest, empowering us to better understand and address the shortcomings of current methods.
The addition of such a scale helps us differentiate between the many works on causal representation learning \citep{scholkopf2021toward,brehmer2022weakly,de2019causal,lippe2022causal,lu2022invariant,ahuja2023interventional,mitrovic2020representation}, and the classical independent components analysis \citep{hyvarinen2013independent,hyvarinen2000independent,zhang2022identifiable,hyvarinen1999nonlinear}.
While the latter helps advance the causal representation learning agenda, it recovers components based on a {\it plausible} causal structure (i.e. the second level on our scale in \cref{fig:scale}), whereas the former recovers components based on a causal structure.

Essentially, through CDL we propose new research to take into account the following practical guidelines:
\begin{enumerate}
    \item {\bf Be clear about causal assumptions.} It is important to leverage the causal knowledge we have \citep{witte2020efficient} but also to take into account knowledge that we don't have; in particular, we need to be careful with using models that ask for total causal knowledge but only give it partially confirmed structures.
    
    \item {\bf Avoid model misspecification by carefully documenting and analysing the parametric setting.} Non-parametric models are powerful allies but are unnecessarily complicated when we already have some parametric knowledge about our setting. Before designing new methods, we encourage a researcher to first analyse the setting in which the new method is to be deployed. Their analysis can vastly reduce model complexity and as a result, allow more powerful inference.
    
    \item {\bf Time deserves special attention from a causal perspective.} As shown in our paper, time is more than simply taking into account an observation's history. Careful analysis is required when time presents new complexities related to confounding.
\end{enumerate}

Furthermore, our framework identifies gaps in current research and proposes avenues for future work, including the need for research on recovering fully known structural equations and reflection on the relevance of current gaps to practical problems. 
We are confident that our framework will make a significant contribution to the advancement of CDL research and its practical applications, paving the way for a more efficient and effective approach to solving complex real-world problems.

\subsubsection*{Acknowledgements}

We truly value the input of the CDL community. We wish to explicitly thank the following community members for their comments (alphabetised by last name):

{\bf Alicia Curth} (University of Cambridge)\\
{\bf Isabelle Guyon} (Google Brain)\\
{\bf Sam Holt} (University of Cambridge)\\
{\bf Alihan H\"uy\"uk} (University of Cambridge)\\
{\bf Trent Kyono} (Meta)\\
{\bf Andrew Rashbass} (University of Cambridge)\\
{\bf Paulius Rauba} (University of Cambridge)\\
{\bf Guillermo Sapiro} (Duke University \& Apple)\\
{\bf Ricardo Silva} (UCL)\\
{\bf Ruibo Tu} (KTH)\\
{\bf Boris van Breugel} (University of Cambridge)\\
{\bf David Watson} (King's College London \& Meta)\\
{\bf Simon Woodhead} (Eedi)\\
{\bf Cheng Zhang} (Microsoft Research)\\
{\bf Kun Zhang} (CMU \& MBZUAI)\\

\bibliography{main}
\bibliographystyle{unsrtnat}

\clearpage
\appendix

\section{Real-world applications} \label{sec:RW}

We discuss the impact of CDL on five different real-world applications: Imputation, Fairness, Generalisation, Treatment effects, and Recommendation. We will examine how CDL can help define interaction effects of missingness and covariates, identify sources of unfairness, constrain methods using a general multi-domain causal structure, learn models that respect identification assumptions, and infer the causal effect of recommendation. 

{\bf Imputation} Imputation is a common problem in many datasets, and missing data can lead to biased solutions. CDL can help define interaction effects of missingness and covariates to improve the accuracy of imputation methods. For example, CDL-based approaches allow deep learning to accurately impute missing values, respecting the causal interaction between missingness indicators and treatment selection \citep{berrevoets2022impute,mayer2020doubly,westreich2015imputation}. CDL-based methods outperform existing imputation techniques in terms of both imputation accuracy {\it and} unbiased estimation from data with missing values. Another approach is to leverage causal knowledge to perform the imputation task directly \citep{squires2022causal,kyono2019improving}.

{\bf Fairness.} CDL can also be used to identify sources of unfairness in machine learning models. In particular, CDL can help identify which features in the data are responsible for observed disparities in outcomes. CDL-based approaches can help mitigate unfairness in machine learning models by learning a representation of the data that is fair concerning a given set of protected attributes and their causal neighbours \citep{loftus2018causal,wu2019pc,van2021decaf,helwegen2020improving}. The authors showed that their method outperforms existing fairness methods in terms of both fairness and accuracy. Other CDL techniques evaluate models post-training using causal evaluation metrics \citep{kusner2017counterfactual}.

{\bf Generalisation.} CDL can help constrain methods using a general multi-domain causal structure, which can improve generalisation performance in new domains. Specifically, causal structures can be used to select models in a new domain as (partial) causal knowledge is assumed invariant \citep{kyono2021exploiting,shi2021invariant,kyono2021selecting}, yielding more robust model selections with increased performance as a result. While it would be beneficial to gather {\it complete} causal knowledge, we can already advance in this application using partial knowledge.

 {\bf Treatment effects.} Perhaps one of the most natural application domains for CDL is estimating causal treatment effects \citep{alaa2017bayesian,shalit2017estimating,curth2021inductive,hassanpour2019counterfactual}. With a long history in statistics, it seems that treatment effect literature is truly benefiting from taking a causal deep learning approach. With applications in donor-organ transplantation \citep{berrevoets2020organite,berrevoets2021learning}, cancer \citep{tabib2020non, basu2014estimating,van2022individual}, and COVID-19 \citep{zame2020machine} and its impact \citep{wozny2022impact}.

{\bf Drug repurposing.} At the intersection of generalisation and treatment effects is drug repurposing, an incredibly exciting area for CDL to have an impact \citep{yang2022machine,liu2020deep}. With applications in COVID-19 \citep{belyaeva2021causal} and dementia \citep{charpignon2022causal,charpignon2021drug}, CDL is already proving to be a useful framework. However, as pointed out in \cref{tab:rw-impact:2}, these high-stakes application domains are a clear example of why we require a careful analysis of the made assumptions.

{\bf Recommendation.} CDL can also be used to infer the causal effect of online store recommendations \citep{sharma2015estimating,wang2020causal,wu2022opportunity}, which is essential for accurate evaluation of recommendation algorithms. Furthermore, \citet{basilico2017deja} illustrate the importance of causality {\it and} time to provide meaningful recommendations, again highlighting the importance of our temporal dimension! Beyond this, \citet{ghazimatin2020prince} show how causality can be used to {\it evaluate} a recommendation system by estimating counterfactual explanations.

\begin{table}[t]
    \centering
    \caption{{\bf Real-world impact.} Some methods in CDL have already impacted the real world. We list some of them in the table above, spanning a few important application areas. In our table, we use the following abbreviations: Unknown (U), Plausible (P), Causal (C), Non-parametric (NP), Noise models (NM), Parametric (Pa.), and Static (St.). These abbreviations correspond with our framework presented in \cref{sec:cartography}.}
    \label{tab:rw-impact:2}
    \begin{tabularx}{\textwidth}{l>{\hsize=1.4\hsize\linewidth=\hsize}X>{\hsize=.6\hsize\linewidth=\hsize}X *{3}{l}}
        \toprule
        {\bf Application} & {\bf How they use causality}  & {\bf Ref.} & \multicolumn{3}{c}{\bf Range of assumptions} \\
        & & & Struct. & Param. & Time \\
        \midrule
        {\it Imputation} & Define missingness and covariates interaction & \citep{morales2022simultaneous,squires2022causal,berrevoets2022impute,westreich2015imputation,mayer2020doubly,kyono2021miracle} & {\footnotesize U $\to$ C} & {\footnotesize NP $\to$ Pa.} & {\footnotesize St.} \\ 
        
        {\it Fairness} & Identify of sources of unfairness & \citep{van2021decaf,loftus2018causal,wu2019pc,kusner2017counterfactual,helwegen2020improving} &
        {\footnotesize U $\to$ C}
        & {\footnotesize NP $\to$ Pa.}  & {\footnotesize St.} \\
        
        {\it Generalisation} & Respect multi-domain causal structure &  \citep{shi2021invariant,kyono2021exploiting,kyono2021selecting} &
        {\footnotesize U $\to$ C}
        & {\footnotesize NP $\to$ NM}  & {\footnotesize St.} \\
        
        {\it Treatment effects} & Respect causal identification assumptions & \citep{alaa2017bayesian,shalit2017estimating,curth2021inductive,hassanpour2019counterfactual,kan2021evaluating, berrevoets2020organite,berrevoets2021learning,tabib2020non,basu2014estimating,van2022individual,zame2020machine,wozny2022impact,bica2020estimating,nazaret2022a} &
        {\footnotesize U $\to$ P}
        & {\footnotesize NP $\to$ Pa.}  & {\footnotesize Both}\\

        {\it Drug repurposing} & Relate multi-modalities using causal structure & \citep{belyaeva2021causal,charpignon2022causal,charpignon2021drug,yang2022machine,liu2020deep} & {\footnotesize U $\to$ C} & {\footnotesize NP $\to$ Pa.} & {\footnotesize Both}. \\

        {\it Recommendation} & Infer causal effect of recommendation & \citep{wang2020causal,basilico2017deja,sharma2015estimating,wu2022opportunity,ghazimatin2020prince} &  
        {\footnotesize U $\to$ P}
        & {\footnotesize NP $\to$ NM} & 
        {\footnotesize Both}
        \\
         
        \bottomrule
    \end{tabularx}
    
\end{table}

We summarise the above in \cref{tab:rw-impact:2}, where we also indicate the range of structural, parametric and temporal assumptions, as per our topographic map. Clearly, from \cref{tab:rw-impact:2} as well as the examples presented in \cref{sec:nav}, it should be clear that we require our framework to truly have detailed discussions and clear adoption of methods going forward.

\section{Categorising CDL methods}\label{sec:nav}

\begin{figure}[t]
    \centering
    \newcommand\Square[1]{+(-#1,-#1) rectangle +(#1,#1)}
    \begin{tikzpicture}[
        grid/.style={gray, dashed},
        scale=.9
    ]
        \node[label={left:Unknown (U)}] (input_knowledge_A) at (0,0) {}; 
        \draw[grid] ($(input_knowledge_A) + (0, .5)$) -- ($(input_knowledge_A) + (4, .5)$);
        \node[label={left:Plausible (P)}] (input_knowledge_U) at ($(input_knowledge_A) + (0, 1)$) {};
        \draw[grid] ($(input_knowledge_U) + (0, .5)$) -- ($(input_knowledge_U) + (4, .5)$);
        \node[label={left:Causal (C)}] (input_knowledge_C) at ($(input_knowledge_U) + (0, 1)$) {};
        
        \node[label=below:(NP)] (input_param_NO) at (.5,-.5) {};
        \node[rotate=-90, anchor=west] at (.5, -1.3) {Non-param.};
        \draw[grid] ($(input_param_NO) + (.5, 0)$) -- ($(input_param_NO) + (.5, 3)$);
        \node[label=below:(NM)] (input_param_1) at ($(input_param_NO) + (1, 0)$) {};
        \node[rotate=-90, anchor=west] at ($(input_param_NO) + (1, -.8)$) {Noise Models};
        \draw[grid] ($(input_param_1) + (.5, 0)$) -- ($(input_param_1) + (.5, 3)$);
        \node[label=below:(Pa.)] (input_param_2) at ($(input_param_1) + (1, 0)$) {};
        \node[rotate=-90, anchor=west] at ($(input_param_1) + (1, -.8)$) {Param.};
        \draw[grid] ($(input_param_2) + (.5, 0)$) -- ($(input_param_2) + (.5, 3)$);
        \node[label={below:{(FK)}}] (input_param_M) at ($(input_param_2) + (1, 0)$) {};
        \node[rotate=-90, anchor=west] at ($(input_param_2) + (1, -.8)$) {Fully known};

        \draw[very thick] ($(input_knowledge_A) - (0, .5)$) -- ($(input_knowledge_C) + (0, .5)$) -- ($(input_knowledge_C) + (4, .5)$) -- ($(input_param_M) + (.5, 0)$) -- cycle;
        
        \node (input_title) at ($(input_param_NO)!.5!(input_param_M) + (0, 3.5)$) {\bf Input};

        
        \node[inner sep=0] (time_start) at ($(input_param_M) + (2, 3.5)$) {};
        \node[inner sep=0] (time_end) at ($(input_param_M) + (2, 1.5)$) {};
        \node[inner sep=0] (static_start) at ($(input_param_M) + (2, .5)$) {};
        \node[inner sep=0] (static_end) at ($(input_param_M) + (2, -1.5)$) {};
        
        \draw[very thick, blue] (time_start) -- (time_end);
        \draw[very thick, blue] (static_start) -- (static_end);
        
        \node at ($(time_start) + (0, .3)$) {\bf \textcolor{blue}{Time}};
        \node at ($(static_start) + (0, .3)$) {\bf \textcolor{blue}{Static}};

        \node[label={left:U}] (repr_knowledge_A) at ($(input_knowledge_A) + (7, 0)$) {}; 
        \draw[grid] ($(repr_knowledge_A) + (0, .5)$) -- ($(repr_knowledge_A) + (4, .5)$);
        \node[label={left:P}] (repr_knowledge_U) at ($(repr_knowledge_A) + (0, 1)$) {};
        \draw[grid] ($(repr_knowledge_U) + (0, .5)$) -- ($(repr_knowledge_U) + (4, .5)$);
        \node[label={left:C}] (repr_knowledge_C) at ($(repr_knowledge_U) + (0, 1)$) {};
        
        \node[label={below:NP}] (repr_param_NO) at ($(repr_knowledge_A) + (.5, -.5)$) {};
        \draw[grid] ($(repr_param_NO) + (.5, 0)$) -- ($(repr_param_NO) + (.5, 3)$);
        \node[label={below:NM}] (repr_param_1) at ($(repr_param_NO) + (1, 0)$) {};
        \draw[grid] ($(repr_param_1) + (.5, 0)$) -- ($(repr_param_1) + (.5, 3)$);
        \node[label={below:Pa.}] (repr_param_2) at ($(repr_param_1) + (1, 0)$) {};
        \draw[grid] ($(repr_param_2) + (.5, 0)$) -- ($(repr_param_2) + (.5, 3)$);
        \node[label={below:FK}] (repr_param_M) at ($(repr_param_2) + (1, 0)$) {};

        \draw[very thick] ($(repr_knowledge_A) - (0, .5)$) -- ($(repr_knowledge_C) + (0, .5)$) -- ($(repr_knowledge_C) + (4, .5)$) -- ($(repr_param_M) + (.5, 0)$) -- cycle;
        
        \node (repr_title) at ($(repr_param_NO)!.5!(repr_param_M) + (0, 3.5)$) {\bf Representation};

        \draw[draw=none, fill=red!40, on layer=back] ($(input_knowledge_A) + (.5, 0)$) \Square{.5};
        
        \draw[draw=none, fill=red!25, on layer=back] ($(input_knowledge_A) + (.5, 0) + 1*(1, 0)$) \Square{.5};
        \draw[draw=none, fill=red!25, on layer=back] ($(input_knowledge_U) + (.5, 0)$) \Square{.5};
        
        \draw[draw=none, fill=red!10, on layer=back] ($(input_knowledge_C) + (.5, 0) + 0*(1, 0)$) \Square{.5};
        \draw[draw=none, fill=red!10, on layer=back] ($(input_knowledge_U) + (.5, 0) + 1*(1, 0)$) \Square{.5};
        \draw[draw=none, fill=red!10, on layer=back] ($(input_knowledge_A) + (.5, 0) + 2*(1, 0)$) \Square{.5};

        \draw[draw=none, fill=red!40, on layer=back] ($(repr_knowledge_C) + (.5, 0)$) \Square{.5};
        
        \draw[draw=none, fill=red!25, on layer=back] ($(repr_knowledge_C) + (.5, 0) + 1*(1, 0)$) \Square{.5};
        \draw[draw=none, fill=red!25, on layer=back] ($(repr_knowledge_U) + (.5, 0)$) \Square{.5};
        
        \draw[draw=none, fill=red!10, on layer=back] ($(repr_knowledge_C) + (.5, 0) + 2*(1, 0)$) \Square{.5};
        \draw[draw=none, fill=red!10, on layer=back] ($(repr_knowledge_U) + (.5, 0) + 1*(1, 0)$) \Square{.5};
        \draw[draw=none, fill=red!10, on layer=back] ($(repr_knowledge_A) + (.5, 0)$) \Square{.5};

        \draw[->,  very thick] ($(input_knowledge_A) + (-1, -1)$)  node[label={left:{\it \textcolor{red}{Goal}}}] {} to[out=0, in=270] ($(input_knowledge_A) + (.5, 0)$);

        \draw[->,  very thick] ($(repr_knowledge_C) + (4.5, -1)$)  node[label={right:{\it \textcolor{red}{Goal}}}] {} to[out=180, in=270] ($(repr_knowledge_C) + (.5, 0)$);

    \end{tikzpicture}
    \caption{{\bf Goals and direction of the field.} We can use the map of causal deep learning to specify the future of our field. Darker shades of red indicate a harder, but more desirable goal. In the case of input, we wish to minimise the required graphical input and made assumptions; with which we hope to achieve a more knowledgeable representation with minimal assumptions.}
    \label{fig:goals}
    \rule{\textwidth}{.5pt}

\end{figure}
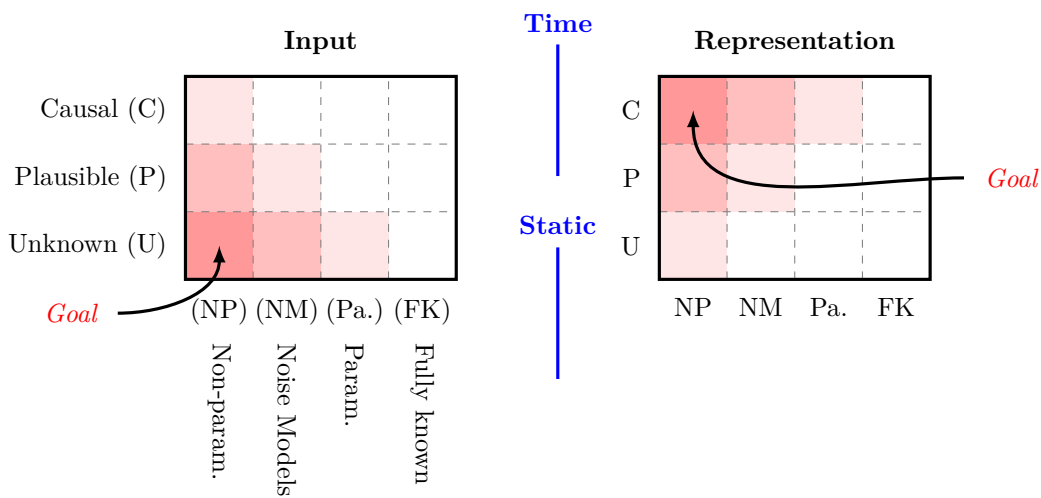

Using the map of causal deep learning we can: (i) categorise and compare literature, as well as (ii) identify some areas that are not well explored. Using these two functionalities, we can perhaps attribute (i) to practitioners, and (ii) to researchers. A practitioner is faced with a problem they wish to solve. Using the map of CDL, the practitioner can map their available data onto the input field and search for a method in the representation field that would solve their needs. Perhaps guided by the required assumptions the practitioner is willing to make, the set of solutions is more digestible than scanning all the literature in CDL. The researcher may use the map differently. Rather than scanning the potential solutions, a researcher can use the map to scan which solutions are still lacking (or more likely, underrepresented).

{\bf Problems and goals.} Learning a structured representation space, while retaining flexible assumptions, is difficult. Consider \cref{fig:goals} where we annotated learning a causally structured representation without any assumptions from arbitrary data with ``\textcolor{red}{\it Goal}''. This learning setup is the most ambitious setup included in our map. It is also long thought to be impossible \citep{glymour2019review,geiger1990logic,meek2013strong,eberhardt2017introduction}.

However, \cref{fig:goals} also shows the relative complexity of other setups. In principle, the less strict assumptions we make, the harder it becomes to narrow down the structure in our representation space. For some problem setups it may be sufficient to learn a flexible structure, which in turn allows for more flexible assumptions on the input. Similarly, if a problem requires a completely identified causal structure, our map shows that one may have make to some strict assumptions on the input. As such, the map of CDL exposes a certain balance between input assumptions and achievable structure.

\subsection{Comparing methods} \label{sec:nav:compare}

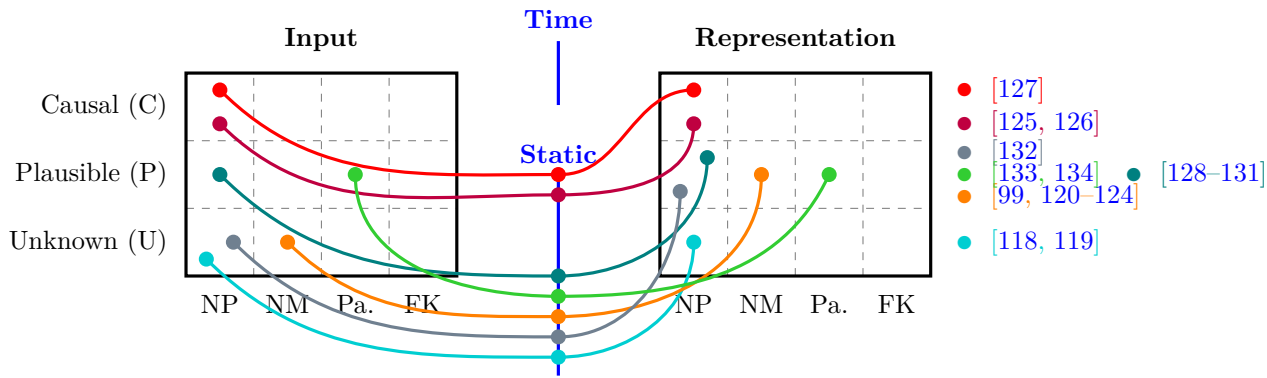
\begin{figure}[t]
    \centering
    \begin{tikzpicture}[
        dot/.style={circle, fill=black, line width=0,  minimum size=2mm, inner sep=0},
        grid/.style={gray, dashed},
        route/.style={very thick, draw opacity=.5},
        scale=.9
    ]
        
        \node[label={left:Unknown (U)}] (input_knowledge_A) at (0,0) {}; 
        \draw[grid] ($(input_knowledge_A) + (0, .5)$) -- ($(input_knowledge_A) + (4, .5)$);
        \node[label={left:Plausible (P)}] (input_knowledge_U) at ($(input_knowledge_A) + (0, 1)$) {};
        \draw[grid] ($(input_knowledge_U) + (0, .5)$) -- ($(input_knowledge_U) + (4, .5)$);
        \node[label={left:Causal (C)}] (input_knowledge_C) at ($(input_knowledge_U) + (0, 1)$) {};
        
        \node[label={below:NP}] (input_param_NO) at (.5,-.5) {};
        \draw[grid] ($(input_param_NO) + (.5, 0)$) -- ($(input_param_NO) + (.5, 3)$);
        \node[label={below:NM}] (input_param_1) at ($(input_param_NO) + (1, 0)$) {};
        \draw[grid] ($(input_param_1) + (.5, 0)$) -- ($(input_param_1) + (.5, 3)$);
        \node[label={below:Pa.}] (input_param_2) at ($(input_param_1) + (1, 0)$) {};
        \draw[grid] ($(input_param_2) + (.5, 0)$) -- ($(input_param_2) + (.5, 3)$);
        \node[label={below:FK}] (input_param_M) at ($(input_param_2) + (1, 0)$) {};

        \draw[very thick] ($(input_knowledge_A) - (0, .5)$) -- ($(input_knowledge_C) + (0, .5)$) -- ($(input_knowledge_C) + (4, .5)$) -- ($(input_param_M) + (.5, 0)$) -- cycle;
        
        \node at ($(input_param_NO)!.5!(input_param_M) + (0, 3.5)$) {\bf Input};

        
        \node[inner sep=0] (time_start) at ($(input_param_M) + (2, 3.5)$) {};
        \node[inner sep=0] (time_end) at ($(input_param_M) + (2, 2.5)$) {};
        \node[inner sep=0] (static_start) at ($(input_param_M) + (2, 1.5)$) {};
        \node[inner sep=0] (static_end) at ($(input_param_M) + (2, -1.5)$) {};
        
        \draw[very thick, blue] (time_start) -- (time_end);
        \draw[very thick, blue] (static_start) -- (static_end);
        
        \node at ($(time_start) + (0, .3)$) {\bf \textcolor{blue}{Time}};
        \node at ($(static_start) + (0, .3)$) {\bf \textcolor{blue}{Static}};

        
        \node (repr_knowledge_A) at ($(input_knowledge_A) + (7, 0)$) {}; 
        \draw[grid] ($(repr_knowledge_A) + (0, .5)$) -- ($(repr_knowledge_A) + (4, .5)$);
        \node (repr_knowledge_U) at ($(repr_knowledge_A) + (0, 1)$) {};
        \draw[grid] ($(repr_knowledge_U) + (0, .5)$) -- ($(repr_knowledge_U) + (4, .5)$);
        \node (repr_knowledge_C) at ($(repr_knowledge_U) + (0, 1)$) {};
        
        \node[label={below:NP}] (repr_param_NO) at ($(repr_knowledge_A) + (.5, -.5)$) {};
        \draw[grid] ($(repr_param_NO) + (.5, 0)$) -- ($(repr_param_NO) + (.5, 3)$);
        \node[label={below:NM}] (repr_param_1) at ($(repr_param_NO) + (1, 0)$) {};
        \draw[grid] ($(repr_param_1) + (.5, 0)$) -- ($(repr_param_1) + (.5, 3)$);
        \node[label={below:Pa.}] (repr_param_2) at ($(repr_param_1) + (1, 0)$) {};
        \draw[grid] ($(repr_param_2) + (.5, 0)$) -- ($(repr_param_2) + (.5, 3)$);
        \node[label={below:FK}] (repr_param_M) at ($(repr_param_2) + (1, 0)$) {};

        \draw[very thick] ($(repr_knowledge_A) - (0, .5)$) -- ($(repr_knowledge_C) + (0, .5)$) -- ($(repr_knowledge_C) + (4, .5)$) -- ($(repr_param_M) + (.5, 0)$) -- cycle;
        
        \node at ($(repr_param_NO)!.5!(repr_param_M) + (0, 3.5)$) {\bf Representation};

        \draw[DarkTurquoise, route] ($(input_knowledge_A) + (.3, -.25) + 0*(1, 0)$)  node[dot, DarkTurquoise] {} to[out=315, in=180] ($(static_start)!.9!(static_end)$) node[dot, DarkTurquoise] {} to[out=0, in=270] ($(repr_knowledge_A) + (.5, 0) + 0*(1, 0)$) node[dot, DarkTurquoise] {};
        \node[label={[text=DarkTurquoise]right:\citep{deng2009imagenet,farago1993strong}}] at ($(repr_knowledge_A) + (4.5, 0)$) {\textcolor{DarkTurquoise}{\dott}};
        
        \draw[orange, route] ($(input_knowledge_A) + (.5, 0) + 1*(1, 0)$) node[dot, orange] {} to[out=315, in=180] ($(static_start)!.7!(static_end)$) node[dot, orange] {} to[out=0, in=270] ($(repr_knowledge_U) + (.5, 0) + 1*(1, 0)$) node[dot, orange] {};
        \node[label={[text=orange]right:\citep{kyono2020castle,kyono2021miracle,yao2021path,hollmann2022tabpfn,fatemi2021slaps,morales2021vicause}}] at ($(repr_knowledge_U) + (4.5, -.33)$) {\textcolor{orange}{\dott}};
        
        \draw[purple, route] ($(input_knowledge_C) + (.5, -.25) + 0*(1, 0)$) node[dot, purple] {} to[out=315, in=180] ($(static_start)!.1!(static_end)$) node[dot, purple] {} to[out=0, in=270] ($(repr_knowledge_C) + (.5, -.25)$) node[dot, purple] {};
        \node[label={[text=purple]right:\citep{russo2022causal,teshima2021incorporating}}] at ($(repr_knowledge_C) + (4.5, -.25)$) {\textcolor{purple}{\dott}};
        
        \draw[red, route] ($(input_knowledge_C) + (.5, .25) + 0*(1, 0)$) node[dot, red] {} to[out=315, in=180] (static_start) node[dot, red] {} to[out=0, in=180] ($(repr_knowledge_C) + (.5, .25)$) node[dot, red] {};
        \node[label={[text=red]right: \citep{kancheti22a}}] at ($(repr_knowledge_C) + (4.5, .25)$) {\textcolor{red}{\dott}};

        \draw[Teal, route] ($(input_knowledge_U) + (.5, 0) + 0*(1, 0)$) node[dot, Teal] {} to[out=315, in=180] ($(static_start)!.5!(static_end)$) node[dot, Teal] {} to[out=0, in=270] ($(repr_knowledge_U) + (.7, .25) + 0*(1, 0)$) node[dot, Teal] {};
        \node[label={[text=Teal]right:\citep{kohavi1996scaling,jiang2008novel,jiang2012improving,jiang2016deep}}] at ($(repr_knowledge_U) + (4.5, 0) + (2.5, 0)$) {\textcolor{Teal}{\dott}};

        \draw[SlateGray, route] ($(input_knowledge_A) + (.7, 0) + 0*(1, 0)$)  node[dot, SlateGray] {} to [out=315, in=180] ($(static_start)!.8!(static_end)$) node[dot, SlateGray] {} to[out=0, in=270] ($(repr_knowledge_U) + (.5, 0) + 0*(1, 0) + (-.2, -.25)$) node[dot, SlateGray] {};
        \node[label={[text=SlateGray]right:\citep{NEURIPS2020_e05c7ba4}}] at ($(repr_knowledge_U) + (4.5, .33)$) {\textcolor{SlateGray}{\dott}};
        
        
        \draw[LimeGreen, route] ($(input_knowledge_U) + (.5, 0) + 2*(1, 0)$) node[dot, LimeGreen] {} to [out=270, in=180] ($(static_start)!.6!(static_end)$) node[dot, LimeGreen] {} to[out=0, in=245]  ($(repr_knowledge_U) + (.5, 0) + 2*(1, 0)$) node[dot, LimeGreen] {};
        \node[label={[text=LimeGreen]right:\citep{duda1973pattern,langley1992analysis}}] at ($(repr_knowledge_U) + (4.5, 0)$) {\textcolor{LimeGreen}{\dott}};

   \end{tikzpicture}
   
   \caption{{\bf Navigating (static) supervised learning.} Here we focus only on methods that share the task of mapping data to a label in a supervised way. We discern between methods based on what input they expect (or assume) and what type of representation they learn {\it before} mapping to a label. Our map presents a straightforward way to categorise methods in causal deep learning. Doing this is useful for practitioners (to identify a suitable method), and to researchers (to identify a potential research gap). Here too: Non-parametric (NP), Noise models (NM), Parametric (Pa.), and Fully known (FK).}
   \label{fig:supervised_learning}
   \rule{\textwidth}{.5pt}
\end{figure}

To show how one can use the map to compare methodologies, we take {\it supervised learning} as an example, presented in \cref{fig:supervised_learning}. Of course, CDL spans more than only supervised learning and so does our map.

To remind ourselves, supervised learning is a problem where we wish to map an input, to a label \citep{murphy2012machine}:
\begin{equation*}
    m: \mathcal{D} \to \mathcal{Y},
\end{equation*}
where $\mathcal{D}$ is once again a dataset, but now including labels: $\{(X_i, Y_i) : i \in [N]\}$ with $X_i \in \mathcal{X}$ and $Y_i \in \mathcal{Y}$. The labels $Y_i \in \mathcal{Y}$ can be anything from a real variable ($\mathcal{Y} = \mathbb{R}$), to a binary label ($\mathcal{Y} = \{0, 1\}$).

As we discussed in \cref{sec:map}, deep learning methods first map the data to a representation before it is mapped to the outcome-label \citep{goodfellow2016deep}: $m: \mathcal{D} \to \mathcal{H} \to \mathcal{Y}$. Here, $\mathcal{H}$ corresponds to the representation as in \cref{eq:cdl_model}. As such, our discussion here does not concern $\mathcal{Y}$, but is instead focused on $\mathcal{H}$ and the structure it may respect.

{\bf Why use CDL for supervised learning?} A fair question indeed. Typically, supervised learning models are evaluated only on the accuracy a model achieves on a hold-out test set. Yet, there is no immediate reason why a causal representation would help in this regard. Rather than maximising accuracy on a hold-out test set, one may resort to CDL to also achieve high accuracy on a subset that is not representative of the training data. Examples such as these include: domain adaptation \citep{zhang2021learning,kyono2021exploiting}, transfer learning \citep{rojas2018invariant,magliacane2017causal}, interpretability \citep{wang2020proactive,moraffah2020causal,kim2019learning,xu2020causality}, or general robustness \citep{wang2020visual,kyono2019improving,muller2021learning,buhlmann2020invariance}.

\subsection{Developing methods} \label{sec:nav:dev}

Let us discuss our map using a non-exhaustive list of some of the key tasks the map of CDL may help a researcher.

{\bf Identification.} A big part in a researcher's workflow is to identify gaps in the literature. For researchers active in supervised learning, our map of CDL can be a useful tool to do exactly this. For example, from \cref{fig:supervised_learning} we learn that the models that take a fully known function as a parametric assumption are not well represented-- they are not represented at all! Of course, this makes sense. If one assumes a fully known model, before commencing the learning task, there is no point learning it anymore. 

Looking further, we find that mapping additive noise models with unknown structure, to additive noise model {\it with} structure actually {\it are} well represented. The reason for this is that recent contributions in differentiable structure learning are a great candidate to regularise models used for other tasks than structure learning. As such, a researcher would learn that there is indeed much competition in this field.

{\bf Navigation and related works.} This brings us to a second task we can employ our map for: building a body of related works. The map will make it easier for researchers to find and learn about related works. Essentially enabling a platform for researchers to share their work with other interested researches. 

Furthermore, the map allows researchers to think critically about the work they present. For example, in \cref{fig:goals} we clearly state that the ``ultimate goal'' to map data to a non-parametric causal structure without making any assumptions is thought to be impossible. Not only will this help researchers to discuss their proposal with a sense of realism, the same can be said about practitioners (and reviewers) alike.

{\bf Data.} Beyond the above, the map also provides guidance for model evaluation. In particular, when evaluating a model, it is important to use data that actually matches the models input assumptions. Similarly for the benchmarks the proposal is compared against. 

For example, if a method assumes a causal graph as input, there should be some assurance that the presented graph is indeed causal. One way to do this is by also provided some empirical evidence that this is the case, such as running clinical trials or assuming interventional data.

\begin{table}[t]
    \centering
    \caption{{\bf Comparing different fields.} \Cref{sec:nav:beyond} compares methods in fields beyond supervised learning (cfr. \cref{sec:nav:compare}). Our table below provides a quick overview of the methods and fields we discuss. Beyond that, we also list the target aimed to learn, as well as the minimum data required to learn that target. We have excluded any additional information (beyond the data) such as parametric or structural assumptions, as this is exactly what the map of CDL can be used for. 
    }
    \label{tab:overview}
    \begin{tabularx}{\linewidth}{X | *{2}{l} | *{2}{l}}
         \toprule
         {\bf Field} & {\bf Data} & {\bf Target} & {\bf References} & {\bf Map} \\
         \midrule
         \textbf{\em Supervised learning (SL)}   & $\{(X_i, Y_i) : i \in [N]\}$ & $\mathcal{X} \to \mathcal{Y}$ & \citep{deng2009imagenet,farago1993strong,kyono2020castle,kyono2021miracle,yao2021path,hollmann2022tabpfn,fatemi2021slaps,morales2021vicause,russo2022causal,teshima2021incorporating,kancheti22a,kohavi1996scaling,jiang2008novel,jiang2012improving,jiang2016deep,NEURIPS2020_e05c7ba4,duda1973pattern,langley1992analysis} & \cref{fig:supervised_learning}\\
         
         \textbf{\em Forecasting/temporal SL} & $\{(X_i, Y_i)_{1:T_i} : i \in [N]\}$ & $\mathcal{T} \times \mathcal{X} \to \mathcal{Y}$& \citep{vaswani2017attention,eichler2017graphical,freeman1983granger,guglielmo97,calderhead2008accelerating,ramsay2007parameter,macdonald2015,wenk19a,dattner2015optimal,oates2014,pfister2018identifying,queen2009,mikkelsen2017learning,pfister2019invariant,Liu_2022_CVPR,sharma2022incorporating,li2020} & \cref{fig:temporal}\\

         
         \textbf{\em Bandits \& RL} & $\{(X_i, A_i, Y_i)_{\pi^b} : i \in [N]\}$ & $\mathcal{X} \times \mathcal{A} \times \mathcal{Y} \to \pi^*$ & \citep{zhang2022causal,lattimore2016causal,lee2018,Lee_Bareinboim_2019,lee2020,nair21a,maiti2022causal,subramanian2022causal,lu2021,bilodeau2022,Berrevoets2022a,sen17a,lu20a,aglietti20a,aglietti2021,tennenholtz21a,sharma2020warm,kroon2020,wang2021,zhang20a,sauter2022a,molina22a,lyle2021resolving,herlau2022reinforcement,Zhu2020Causal,dasgupta2019causal,seitzer2021,yang2021causal,wang2021ordering,he2022causal,lu22a,zhu2022causal,shi2022dynamic,gasse2021causal,mutti2022provably} & \cref{fig:bandits}\\


         \textbf{\em Generative modelling} & $\{(X_i) : i \in [N]\}$ & $\mathcal{X} \to \mathcal{X}$ & \citep{wen2021causal,kocaoglu2018causalgan,sauer2021counterfactual,depeng,shen2022weakly,van2021decaf,cinquini2021,goudet2017causal,Mao_2021_CVPR,zhang2022cmgan,li2022gflowcausal,liu2023goggle} & \cref{fig:generative}\\

         \bottomrule
    \end{tabularx}
    
\end{table}

\subsection{Using methods} \label{sec:nav:use}

{\bf Navigate.} While the researcher's perspective is mostly based around identifying methods that {\it don't} exist, the practitioner's is the opposite. When faced with a certain problem, the practitioner may use the map to identify which methods assume the input the practitioner has available, while also solving the problem they wish to solve. 

Finding a suitable method can be extremely challenging for practitioners as there are new methods proposed almost daily. This latter point is in fact one of the prime reasons why we propose the map in the first place. 

{\bf Validation and education.} On the other hand, it may be that the practitioner wishes to solve a problem that is not yet solved before (stumbling upon a gap in the literature). Or, more strikingly, they are faced with a problem that is simply impossible to solve.

Especially the latter could be an interesting use of our map: education. We strongly believe that the map of CDL can form a bridge between researches and practitioners. Educating which problems {\it can} be solved (which is communicated from research to practice), and which problems {\it have to be} solved (which is in turn communicated from practice to research). 

As we have observed in \cref{fig:supervised_learning}, we find that there is a heavy focus on learning structure from nothing in the additive noise setting. But, perhaps this is not interesting from a practical point of view as it could be that practitioners are well aware of some conditional independence, or some parametric shape of their environment. Having clear communication between both parties may avoid researchers  to waste valuable resources and time to solve problems that need not be solved.

\begin{figure}[t]
    \centering
    \begin{tikzpicture}[
        dot/.style={circle, fill=black, line width=0,  minimum size=2mm, inner sep=0},
        grid/.style={gray, dashed},
        route/.style={very thick, draw opacity=.5},
        scale=.9
    ]
        
        \node[label={left:Unknown (U)}] (input_knowledge_A) at (0,0) {}; 
        \draw[grid] ($(input_knowledge_A) + (0, .5)$) -- ($(input_knowledge_A) + (4, .5)$);
        \node[label={left:Plausible (P)}] (input_knowledge_U) at ($(input_knowledge_A) + (0, 1)$) {};
        \draw[grid] ($(input_knowledge_U) + (0, .5)$) -- ($(input_knowledge_U) + (4, .5)$);
        \node[label={left:Causal (C)}] (input_knowledge_C) at ($(input_knowledge_U) + (0, 1)$) {};
        
        \node[label={below:NP}] (input_param_NO) at (.5,-.5) {};
        \draw[grid] ($(input_param_NO) + (.5, 0)$) -- ($(input_param_NO) + (.5, 3)$);
        \node[label={below:NM}] (input_param_1) at ($(input_param_NO) + (1, 0)$) {};
        \draw[grid] ($(input_param_1) + (.5, 0)$) -- ($(input_param_1) + (.5, 3)$);
        \node[label={below:Pa.}] (input_param_2) at ($(input_param_1) + (1, 0)$) {};
        \draw[grid] ($(input_param_2) + (.5, 0)$) -- ($(input_param_2) + (.5, 3)$);
        \node[label={below:FK}] (input_param_M) at ($(input_param_2) + (1, 0)$) {};

        \draw[very thick] ($(input_knowledge_A) - (0, .5)$) -- ($(input_knowledge_C) + (0, .5)$) -- ($(input_knowledge_C) + (4, .5)$) -- ($(input_param_M) + (.5, 0)$) -- cycle;
        
        \node at ($(input_param_NO)!.5!(input_param_M) + (0, 3.5)$) {\bf Input};

        
        \node[inner sep=0] (time_start) at ($(input_param_M) + (2, 3.5)$) {};
        \node[inner sep=0] (time_end) at ($(input_param_M) + (2, .5)$) {};
        \node[inner sep=0] (static_start) at ($(input_param_M) + (2, -.5)$) {};
        \node[inner sep=0] (static_end) at ($(input_param_M) + (2, -1.5)$) {};
        
        \draw[very thick, blue] (time_start) -- (time_end);
        \draw[very thick, blue] (static_start) -- (static_end);
        
        \node at ($(time_start) + (0, .3)$) {\bf \textcolor{blue}{Time}};
        \node at ($(static_start) + (0, .3)$) {\bf \textcolor{blue}{Static}};

        
        \node (repr_knowledge_A) at ($(input_knowledge_A) + (7, 0)$) {}; 
        \draw[grid] ($(repr_knowledge_A) + (0, .5)$) -- ($(repr_knowledge_A) + (4, .5)$);
        \node (repr_knowledge_U) at ($(repr_knowledge_A) + (0, 1)$) {};
        \draw[grid] ($(repr_knowledge_U) + (0, .5)$) -- ($(repr_knowledge_U) + (4, .5)$);
        \node (repr_knowledge_C) at ($(repr_knowledge_U) + (0, 1)$) {};
        
        \node[label={below:NP}] (repr_param_NO) at ($(repr_knowledge_A) + (.5, -.5)$) {};
        \draw[grid] ($(repr_param_NO) + (.5, 0)$) -- ($(repr_param_NO) + (.5, 3)$);
        \node[label={below:NM}] (repr_param_1) at ($(repr_param_NO) + (1, 0)$) {};
        \draw[grid] ($(repr_param_1) + (.5, 0)$) -- ($(repr_param_1) + (.5, 3)$);
        \node[label={below:Pa.}] (repr_param_2) at ($(repr_param_1) + (1, 0)$) {};
        \draw[grid] ($(repr_param_2) + (.5, 0)$) -- ($(repr_param_2) + (.5, 3)$);
        \node[label={below:FK}] (repr_param_M) at ($(repr_param_2) + (1, 0)$) {};

        \draw[very thick] ($(repr_knowledge_A) - (0, .5)$) -- ($(repr_knowledge_C) + (0, .5)$) -- ($(repr_knowledge_C) + (4, .5)$) -- ($(repr_param_M) + (.5, 0)$) -- cycle;
        
        \node at ($(repr_param_NO)!.5!(repr_param_M) + (0, 3.5)$) {\bf Representation};

        \draw[DarkTurquoise, route] ($(input_knowledge_C) + (.5, 0) + 1*(1, 0)$)  node[dot, DarkTurquoise] {} to[out=27, in=180] ($(time_start)!.1!(time_end)$) node[dot, DarkTurquoise] {} to[out=0, in=135] ($(repr_knowledge_C) + (.25, -.25) + 1*(1, 0)$) node[dot, DarkTurquoise] {};
        \node[label={[text=DarkTurquoise]right:\citep{eichler2017graphical}}] at ($(repr_knowledge_C) + (4.5, -.25)$) {\textcolor{DarkTurquoise}{\dott}};

        \draw[orange, route] ($(input_knowledge_C) + (.5, 0) + 2*(1, 0)$) node[dot, orange] {} to[out=315, in=180] ($(time_start)!.5!(time_end)$) node[dot, orange] {} to[out=0, in=225] ($(repr_knowledge_C) + (.25, -.25) + 2*(1, 0)$) node[dot, orange] {};
        \node[label={[text=orange]right:\citep{freeman1983granger,guglielmo97}}] at ($(repr_knowledge_C) + (6, -.25)$) {\textcolor{orange}{\dott}};
        
        \draw[purple, route] ($(input_knowledge_A) + (.25, -.25) + 1*(1, 0)$) node[dot, purple] {} to[out=45, in=180] ($(time_start)!.6!(time_end)$) node[dot, purple] {} to[out=0, in=135] ($(repr_knowledge_A) + (.5, 0) + 1*(1, 0)$) node[dot, purple] {};
        \node[label={[text=purple]right:\citep{calderhead2008accelerating,ramsay2007parameter,macdonald2015,wenk19a}}] at ($(repr_knowledge_A) + (4.5, -.25)$) {\textcolor{purple}{\dott}};
        
        \draw[red, route] ($(input_knowledge_A) + (.75, -.25) +  2*(1, 0)$) node[dot, red] {} to[out=45, in=180] ($(time_start)!.7!(time_end)$) node[dot, red] {} to[out=0, in=135] ($(repr_knowledge_A) + (.5, 0) +  2*(1, 0)$) node[dot, red] {};
        \node[label={[text=red]right: \citep{dattner2015optimal}}] at ($(repr_knowledge_A) + (6, .25)$) {\textcolor{red}{\dott}};

        \draw[Teal, route] ($(input_knowledge_A) + (.25, .25) + 1*(1, 0)$) node[dot, Teal] {} to[out=45, in=180] ($(time_start)!.2!(time_end)$) node[dot, Teal] {} to[out=0, in=180] ($(repr_knowledge_C) + (.7, .25) + 1*(1, 0)$) node[dot, Teal] {};
        \node[label={[text=Teal]right:\citep{oates2014,pfister2018identifying}}] at ($(repr_knowledge_C) + (4.5, .25)$) {\textcolor{Teal}{\dott}};

        \draw[SlateGray, route] ($(input_knowledge_A) + (.7, 0) + 1*(1, 0)$)  node[dot, SlateGray] {} to [out=45, in=180] ($(time_start)!.4!(time_end)$) node[dot, SlateGray] {} to[out=0, in=135] ($(repr_knowledge_U) + (.5, 0) + 1*(1, 0)$) node[dot, SlateGray] {};
        \node[label={[text=SlateGray]right:\citep{mikkelsen2017learning}}] at ($(repr_knowledge_U) + (4.5, -.25)$) {\textcolor{SlateGray}{\dott}};

        \draw[LimeGreen, route] ($(input_knowledge_A) + (.25, .25) + 2*(1, 0)$) node[dot, LimeGreen] {} to [out=45, in=180] ($(time_start)!.3!(time_end)$) node[dot, LimeGreen] {} to[out=0, in=200]  ($(repr_knowledge_C) + (.75, .25) + 2*(1, 0)$) node[dot, LimeGreen] {};
        \node[label={[text=LimeGreen]right:\citep{pfister2019invariant,queen2009}}] at ($(repr_knowledge_C) + (6.5, .25)$) {\textcolor{LimeGreen}{\dott}};

        \draw[Goldenrod, route] ($(input_knowledge_A) + (.25, .25) + 0*(1, 0)$)  node[dot, Goldenrod] {} to [out=45, in=180] ($(time_start)!.8!(time_end)$) node[dot, Goldenrod] {} to[out=0, in=200]  ($(repr_knowledge_U) + (.75, .25) + 0*(1, 0)$) node[dot, Goldenrod] {};
        \node[label={[text=Goldenrod]right:\citep{Liu_2022_CVPR}}] at ($(repr_knowledge_U) + (4.5, .25)$) {\textcolor{Goldenrod}{\dott}};

        \draw[blue, route] ($(input_knowledge_A) + (.75, -.25) + 0*(1, 0)$)  node[dot, blue] {} to [out=315, in=180] ($(time_start)!.9!(time_end)$) node[dot, blue] {} to[out=0, in=155]  ($(repr_knowledge_A) + (.5, 0) + 0*(1, 0)$) node[dot, blue] {};
        \node[label={[text=blue]right:\citep{vaswani2017attention}}] at ($(repr_knowledge_A) + (4.5, .25)$) {\textcolor{blue}{\dott}};
        
        \draw[Violet, route] ($(input_knowledge_C) + (.5, 0) + 0*(1, 0)$)  node[dot, Violet] {} to [out=35, in=180] (time_start) node[dot, Violet] {} to[out=0, in=135]  ($(repr_knowledge_C) + (.5, 0) + 0*(1, 0)$) node[dot, Violet] {};
        \node[label={[text=Violet]right:\citep{sharma2022incorporating}}] at ($(repr_knowledge_C) + (8, -.25)$) {\textcolor{Violet}{\dott}};

        \draw[Aquamarine, route] ($(input_knowledge_U) + (.5, 0) + 0*(1, 0)$)  node[dot, Aquamarine] {} to [out=35, in=180] (time_end) node[dot, Aquamarine] {} to[out=0, in=225]  ($(repr_knowledge_U) + (.25, -.25) + 0*(1, 0)$) node[dot, Aquamarine] {};
        \node[label={[text=Aquamarine]right:\citep{li2020}}] at ($(repr_knowledge_U) + (6, -.25)$) {\textcolor{Aquamarine}{\dott}};

   \end{tikzpicture}
   
   \caption{{\bf Navigating (temporal) supervised learning and forecasting.} Here we focus on methods that solve a prediction task in the temporal domain. As in \cref{fig:supervised_learning} we discern between methods based on what input they expect (or assume) and what type of representation they learn {\it before} mapping to a label. Here too: Non-parametric (NP), Noise models (NM), Parametric (Pa.), and Fully known (FK).}
   \label{fig:temporal}
   \rule{\textwidth}{.5pt}
\end{figure}
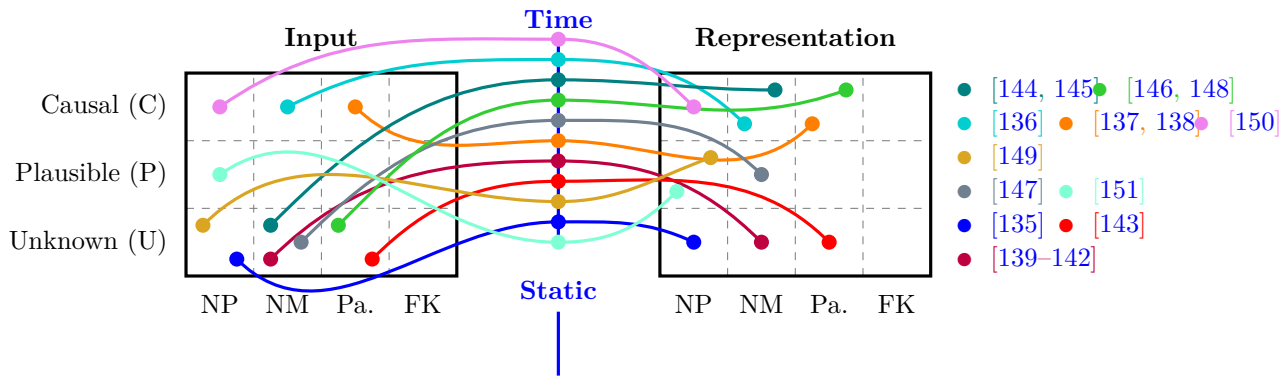

\subsection{Beyond supervised learning} \label{sec:nav:beyond}

Having explained how our map can categorise methods within the supervised learning problem (and what benefits this yields in \cref{sec:nav:compare,sec:nav:dev,sec:nav:use}), let us now consider other machine learning problems. We have listed these in \cref{tab:overview} where we-- like we have for supervised learning in \cref{sec:nav:compare} --present: the minimal data required to solve the problem, the problem's target, and provide a link to the map specific to the problem as well as the references they categorise. Naturally, more references and problems can be added, but we believe \cref{tab:overview} spans a broad range of the most researched ML problems. Thereby showing the versatility of the map of CDL.

\subsubsection{Forecasting and temporal supervised learning}

Perhaps a logical {\it next step} beyond supervised learning is supervised learning in a dynamic setting. In particular, we will now categorise methods that forecast a prediction multiple timesteps ahead, both in discrete as in continuous time. As with static supervised learning, we assume each method has access to at least a dataset comprising features and labels. The difference is, though, that the features (and potentially the label) may be observed more than once. For example, for one patient, we may observe their blood-pressure at multiple timepoints (with a possibly changing blood-pressure). An overview of methods in this category can be found in \cref{fig:temporal}.

Strikingly, when we compare the static setting (\cref{fig:supervised_learning}), with the temporal setting (\cref{fig:temporal}), we see that the static setting seems to be much more focused on non-parametric estimation than the temporal setting. Recall that, this difference is situated entirely on the parametric scale, which was not considered prior to the introduction of our map of CDL! If we were to only consider the structural scale (as one would when categorising according the ladder of causation in \cref{fig:PCH:ladder}), it would appear that the static setting and the temporal setting were quite similar in focus, i.e. papers would be similarly distributed over the unknown, plausible, and causal categories in \cref{fig:scale}.

\subsubsection{Bandit algorithms and reinforcement learning}

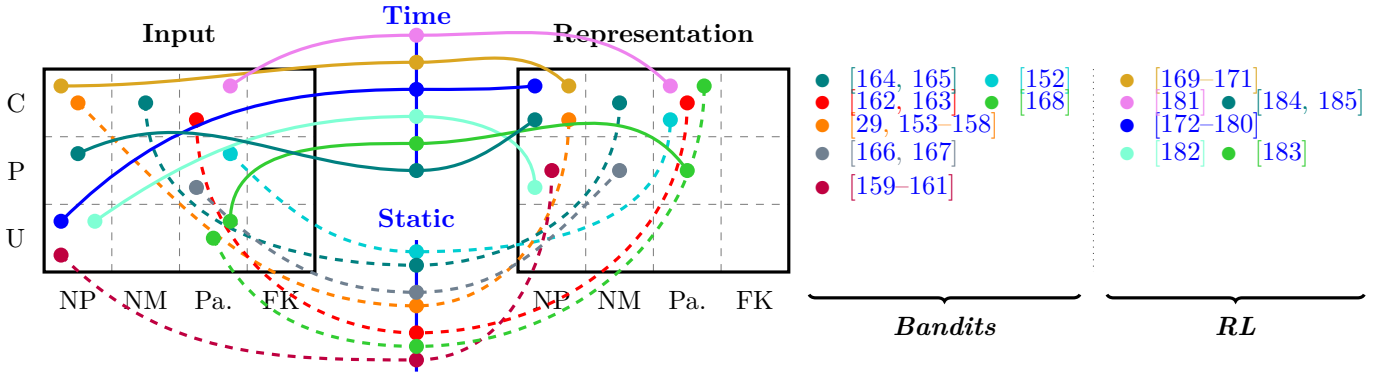
\begin{figure}[t]
    \centering
    \hspace*{-1.2cm}  
    \begin{tikzpicture}[
        dot/.style={circle, fill=black, line width=0,  minimum size=2mm, inner sep=0},
        grid/.style={gray, dashed},
        route/.style={very thick, draw opacity=.5},
        bandit/.style={dashed},
        scale=.9
    ]
        
        \node[label={left:U}] (input_knowledge_A) at (0,0) {}; 
        \draw[grid] ($(input_knowledge_A) + (0, .5)$) -- ($(input_knowledge_A) + (4, .5)$);
        \node[label={left:P}] (input_knowledge_U) at ($(input_knowledge_A) + (0, 1)$) {};
        \draw[grid] ($(input_knowledge_U) + (0, .5)$) -- ($(input_knowledge_U) + (4, .5)$);
        \node[label={left:C}] (input_knowledge_C) at ($(input_knowledge_U) + (0, 1)$) {};
        
        \node[label={below:NP}] (input_param_NO) at (.5,-.5) {};
        \draw[grid] ($(input_param_NO) + (.5, 0)$) -- ($(input_param_NO) + (.5, 3)$);
        \node[label={below:NM}] (input_param_1) at ($(input_param_NO) + (1, 0)$) {};
        \draw[grid] ($(input_param_1) + (.5, 0)$) -- ($(input_param_1) + (.5, 3)$);
        \node[label={below:Pa.}] (input_param_2) at ($(input_param_1) + (1, 0)$) {};
        \draw[grid] ($(input_param_2) + (.5, 0)$) -- ($(input_param_2) + (.5, 3)$);
        \node[label={below:FK}] (input_param_M) at ($(input_param_2) + (1, 0)$) {};

        \draw[very thick] ($(input_knowledge_A) - (0, .5)$) -- ($(input_knowledge_C) + (0, .5)$) -- ($(input_knowledge_C) + (4, .5)$) -- ($(input_param_M) + (.5, 0)$) -- cycle;
        
        \node at ($(input_param_NO)!.5!(input_param_M) + (0, 3.5)$) {\bf Input};

        
        \node[inner sep=0] (time_start) at ($(input_param_M) + (2, 3.5)$) {};
        \node[inner sep=0] (time_end) at ($(input_param_M) + (2, 1.5)$) {};
        \node[inner sep=0] (static_start) at ($(input_param_M) + (2, .5)$) {};
        \node[inner sep=0] (static_end) at ($(input_param_M) + (2, -1.5)$) {};
        
        \draw[very thick, blue] (time_start) -- (time_end);
        \draw[very thick, blue] (static_start) -- (static_end);
        
        \node at ($(time_start) + (0, .3)$) {\bf \textcolor{blue}{Time}};
        \node at ($(static_start) + (0, .3)$) {\bf \textcolor{blue}{Static}};

        
        \node (repr_knowledge_A) at ($(input_knowledge_A) + (7, 0)$) {}; 
        \draw[grid] ($(repr_knowledge_A) + (0, .5)$) -- ($(repr_knowledge_A) + (4, .5)$);
        \node (repr_knowledge_U) at ($(repr_knowledge_A) + (0, 1)$) {};
        \draw[grid] ($(repr_knowledge_U) + (0, .5)$) -- ($(repr_knowledge_U) + (4, .5)$);
        \node (repr_knowledge_C) at ($(repr_knowledge_U) + (0, 1)$) {};
        
        \node[label={below:NP}] (repr_param_NO) at ($(repr_knowledge_A) + (.5, -.5)$) {};
        \draw[grid] ($(repr_param_NO) + (.5, 0)$) -- ($(repr_param_NO) + (.5, 3)$);
        \node[label={below:NM}] (repr_param_1) at ($(repr_param_NO) + (1, 0)$) {};
        \draw[grid] ($(repr_param_1) + (.5, 0)$) -- ($(repr_param_1) + (.5, 3)$);
        \node[label={below:Pa.}] (repr_param_2) at ($(repr_param_1) + (1, 0)$) {};
        \draw[grid] ($(repr_param_2) + (.5, 0)$) -- ($(repr_param_2) + (.5, 3)$);
        \node[label={below:FK}] (repr_param_M) at ($(repr_param_2) + (1, 0)$) {};

        \draw[very thick] ($(repr_knowledge_A) - (0, .5)$) -- ($(repr_knowledge_C) + (0, .5)$) -- ($(repr_knowledge_C) + (4, .5)$) -- ($(repr_param_M) + (.5, 0)$) -- cycle;
        
        \node at ($(repr_param_NO)!.5!(repr_param_M) + (0, 3.5)$) {\bf Representation};


        \draw[DarkTurquoise, route, bandit] ($(input_knowledge_U) + (.75, .25) + 2*(1, 0)$)  node[dot, DarkTurquoise] {} to[out=315, in=180] ($(static_start)!.1!(static_end)$) node[dot, DarkTurquoise] {} to[out=0, in=270 ] ($(repr_knowledge_C) + (.25, -.25) + 2*(1, 0)$) node[dot, DarkTurquoise] {};
        \node[label={[text=DarkTurquoise]right:\citep{zhang2022causal}}] at ($(repr_knowledge_C) + (7, .33)$) {\textcolor{DarkTurquoise}{\dott}};

        \draw[orange, route, bandit] ($(input_knowledge_C) + (.5, 0) + 0*(1, 0)$) node[dot, orange] {} to[out=315, in=180] ($(static_start)!.5!(static_end)$) node[dot, orange] {} to[out=0, in=270] ($(repr_knowledge_C) + (.75, -.25) + 0*(1, 0)$) node[dot, orange] {};
        \node[label={[text=orange]right:\citep{lattimore2016causal,lee2018,Lee_Bareinboim_2019,lee2020,nair21a,maiti2022causal,subramanian2022causal}}] at ($(repr_knowledge_C) + (4.5, -.33)$) {\textcolor{orange}{\dott}};

        \draw[purple, route, bandit] ($(input_knowledge_A) + (.25, -.25) + 0*(1, 0)$) node[dot, purple] {} to[out=315, in=180] ($(static_start)!.9!(static_end)$) node[dot, purple] {} to[out=0, in=270] ($(repr_knowledge_U) + (.5, 0) + 0*(1, 0)$) node[dot, purple] {};
        \node[label={[text=purple]right:\citep{lu2021,bilodeau2022,Berrevoets2022a}}] at ($(repr_knowledge_U) + (4.5, -.25)$) {\textcolor{purple}{\dott}};
       
        \draw[red, route, bandit] ($(input_knowledge_C) + (.25, -.25) +  2*(1, 0)$) node[dot, red] {} to[out=270, in=180] ($(static_start)!.7!(static_end)$) node[dot, red] {} to[out=0, in=270] ($(repr_knowledge_C) + (.5, 0) +  2*(1, 0)$) node[dot, red] {};
        \node[label={[text=red]right: \citep{sen17a,lu20a}}] at ($(repr_knowledge_C) + (4.5, 0)$) {\textcolor{red}{\dott}};
        
        \draw[Teal, route, bandit] ($(input_knowledge_C) + (.5, 0) + 1*(1, 0)$) node[dot, Teal] {} to[out=270, in=180] ($(static_start)!.2!(static_end)$) node[dot, Teal] {} to[out=0, in=270] ($(repr_knowledge_C) + (.5, 0) + 1*(1, 0)$) node[dot, Teal] {};
        \node[label={[text=Teal]right:\citep{aglietti20a,aglietti2021}}] at ($(repr_knowledge_C) + (4.5, .33)$) {\textcolor{Teal}{\dott}};
        
        \draw[SlateGray, route, bandit] ($(input_knowledge_U) + (.25, -.25) + 2*(1, 0)$)  node[dot, SlateGray] {} to [out=315, in=180] ($(static_start)!.4!(static_end)$) node[dot, SlateGray] {} to[out=0, in=225] ($(repr_knowledge_U) + (.5, 0) + 1*(1, 0)$) node[dot, SlateGray] {};
        \node[label={[text=SlateGray]right:\citep{tennenholtz21a,sharma2020warm}}] at ($(repr_knowledge_U) + (4.5, .25)$) {\textcolor{SlateGray}{\dott}};
        
        \draw[LimeGreen, route, bandit] ($(input_knowledge_A) + (.5, 0) + 2*(1, 0)$) node[dot, LimeGreen] {} to [out=295, in=180] ($(static_start)!.8!(static_end)$) node[dot, LimeGreen] {} to[out=0, in=270]  ($(repr_knowledge_C) + (.75, .25) + 2*(1, 0)$) node[dot, LimeGreen] {};
        \node[label={[text=LimeGreen]right:\citep{kroon2020}}] at ($(repr_knowledge_C) + (7, 0)$) {\textcolor{LimeGreen}{\dott}};

        \draw[dotted, black!70] ($(repr_knowledge_C) + (8.5, .5)$) -- ($(repr_knowledge_A) + (8.5, -.5)$);


        \draw[Goldenrod, route] ($(input_knowledge_C) + (.25, .25) + 0*(1, 0)$)  node[dot, Goldenrod] {} to [out=0, in=180] ($(time_start)!.2!(time_end)$) node[dot, Goldenrod] {} to[out=0, in=135]  ($(repr_knowledge_C) + (.75, .25) + 0*(1, 0)$) node[dot, Goldenrod] {};
        \node[label={[text=Goldenrod]right:\citep{wang2021,zhang20a,sauter2022a}}] at ($(repr_knowledge_C) + (9, .33)$) {\textcolor{Goldenrod}{\dott}};
        
        \draw[blue, route] ($(input_knowledge_A) + (.25, .25) + 0*(1, 0)$)  node[dot, blue] {} to [out=45, in=180] ($(time_start)!.4!(time_end)$) node[dot, blue] {} to[out=0, in=185]  ($(repr_knowledge_C) + (.25, .25) + 0*(1, 0)$) node[dot, blue] {};
        \node[label={[text=blue]right:\citep{molina22a,lyle2021resolving,herlau2022reinforcement,Zhu2020Causal,dasgupta2019causal,seitzer2021,yang2021causal,wang2021ordering,he2022causal}}] at ($(repr_knowledge_C) + (9, -.33)$) {\textcolor{blue}{\dott}};

        \draw[Violet, route] ($(input_knowledge_C) + (.75, .25) + 2*(1, 0)$)  node[dot, Violet] {} to [out=35, in=180] (time_start) node[dot, Violet] {} to[out=0, in=135]  ($(repr_knowledge_C) + (.25, .25) + 2*(1, 0)$) node[dot, Violet] {};
        \node[label={[text=Violet]right:\citep{lu22a}}] at ($(repr_knowledge_C) + (9, 0)$) {\textcolor{Violet}{\dott}};
        
        \draw[Aquamarine, route] ($(input_knowledge_A) + (.75, .25) + 0*(1, 0)$)  node[dot, Aquamarine] {} to [out=35, in=180] ($(time_start)!.6!(time_end)$) node[dot, Aquamarine] {} to[out=0, in=90]  ($(repr_knowledge_U) + (.25, -.25) + 0*(1, 0)$) node[dot, Aquamarine] {};
        \node[label={[text=Aquamarine]right:\citep{zhu2022causal}}] at ($(repr_knowledge_U) + (9, .25)$) {\textcolor{Aquamarine}{\dott}};

        \draw[LimeGreen, route] ($(input_knowledge_A) + (.75, .25) + 2*(1, 0)$)  node[dot, LimeGreen] {} to [out=90, in=180] ($(time_start)!.8!(time_end)$) node[dot, LimeGreen] {} to[out=0, in=125]  ($(repr_knowledge_U) + (.5, 0) + 2*(1, 0)$) node[dot, LimeGreen] {};
        \node[label={[text=LimeGreen]right:\citep{shi2022dynamic}}] at ($(repr_knowledge_U) + (10.5, .25)$) {\textcolor{LimeGreen}{\dott}};

        \draw[Teal, route] ($(input_knowledge_U) + (.5, .25) + 0*(1, 0)$)  node[dot, Teal] {} to [out=25, in=180] ($(time_start)!1!(time_end)$) node[dot, Teal] {} to[out=0, in=215]  ($(repr_knowledge_C) + (.25, -.25) + 0*(1, 0)$) node[dot, Teal] {};
        \node[label={[text=Teal]right:\citep{gasse2021causal,mutti2022provably}}] at ($(repr_knowledge_C) + (10.5, .0)$) {\textcolor{Teal}{\dott}};

        \draw [
            very thick,
            decoration={
                brace,
                mirror,
                raise=-.15cm,
            },
            decorate
        ] ($(repr_knowledge_A) + (4.3, -1)$) -- ($(repr_knowledge_A) + (8.3, -1)$) node [pos=0.5,anchor=center, below] {\textbf{\em Bandits}};

        \draw [
            very thick,
            decoration={
                brace,
                mirror,
                raise=-.15cm,
            },
            decorate
        ] ($(repr_knowledge_A) + (8.7, -1)$) -- ($(repr_knowledge_A) + (12.5, -1)$) node [pos=0.5,anchor=center, below] {\textbf{\em RL}};

   \end{tikzpicture}
   
   \caption{{\bf Navigating bandit algorithms and reinforcement learning (RL).} Here we focus on methods that have interaction in the system. In particular, we consider bandit algorithms (where a sequence has no meaning), and RL (where sequence {\it has} meaning). As in \cref{fig:supervised_learning,fig:temporal} we discern between methods based on what input they expect (or assume) and what type of representation they learn {\it before} mapping to a label. Here too: Non-parametric (NP), Noise models (NM), Parametric (Pa.), and Fully known (FK); and Unknown (U), Plausible (P), and Causal (C).}
   \label{fig:bandits}
   \rule{\textwidth}{.5pt}
\end{figure}

In \cref{fig:bandits} we have classified both bandit algorithms and reinforcement learning (RL) algorithms. The reason we categorise them together, is that they only differ in dependence across time (or sequential decisions). As such, they are easily differentiated using our Time/Static axis between input and representation. However, to clarify things further, we use dashed lines to represent bandit algorithms and full lines to represent RL algorithms.

\textbf{\em Bandits.} Let us first discuss the bandit side. From \cref{fig:bandits} we can make a few observations. A first is that bandit algorithms seem to be very explicit when it comes to making structural assumptions. We believe a reason for this is that causal bandits tend to come from the PCH community. In contrast, it is not very common to make explicit the assumptions made on a parametric level. That is, most bandit papers are written in a non-parametric setting, or a linear setting, with the exception of a few algorithms situated in Bayesian optimisation (using Gaussian processes). Essentially, these papers provide regret analysis completely agnostic of the type of estimator used infer reward. Furthermore, only a few papers make {\it structural transitions}, where the algorithm aims to learn causal information, rather than simply assuming access to it.

\textbf{\em RL.} A second in \cref{fig:bandits} are the reinforcement learning algorithms. Combining bandits and RL in one figure shows some interesting differences and equivalencies. First, it seems that-- like in bandits --RL is also focused on non-parametric situations. The reason for this is a little different though. While research on bandits seem to provide algorithms which are agnostic to estimation, RL seems to stem more from a deep learning community; hence most papers in RL rely on neural networks to approximate the reward (based on, for example, \citet{mnih2015human}). 

Another important difference is the amount of RL papers that seem to succeed in a {\it structural transition} from unknown structure to causal structure, even in the non-parametric setting! Compare this with \cref{fig:supervised_learning,fig:temporal} where {\it none} of the algorithms achief this. Why is this? Contrasting supervised learning, bandits and RL algorithms are allowed to perform {\it interventions} in the environment, whereas supervised learning algorithms can only rely on observational data. These experiments are of course hugely informative when inferring causality. 

This brings us at an import point when considering \cref{fig:bandits}: we have only included papers that make an explicit point about causality in their investigation. However, recent work has shown that there may be an even deeper connection between causality and bandits and RL \citep{bannon2020}. In fact, some results seem to indicate that some models may even exhibit causal knowledge without being trained for it explicitly \citep{dasgupta2019causal}. We hope that these results encourage RL-researchers to consider causality in their papers.

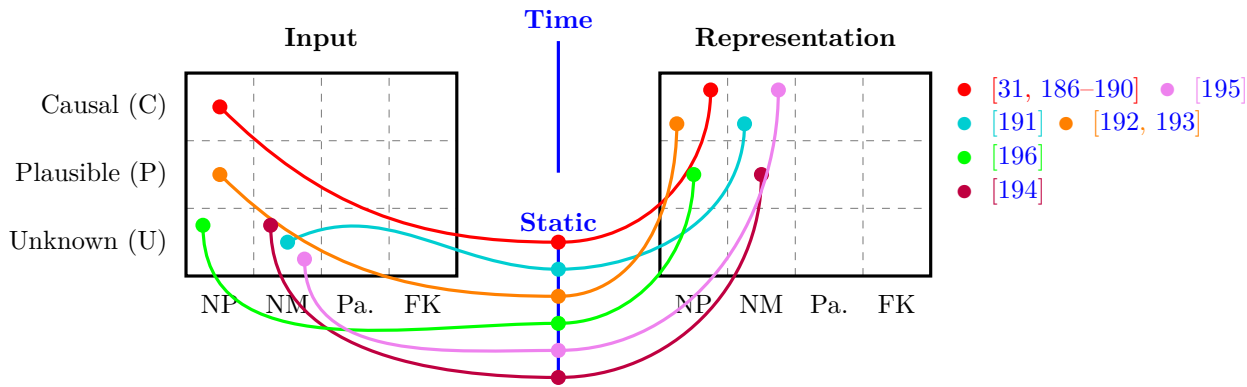
\begin{figure}[t]
    \centering
    \begin{tikzpicture}[
        dot/.style={circle, fill=black, line width=0,  minimum size=2mm, inner sep=0},
        grid/.style={gray, dashed},
        route/.style={very thick, draw opacity=.5},
        scale=.9
    ]
        
        \node[label={left:Unknown (U)}] (input_knowledge_A) at (0,0) {}; 
        \draw[grid] ($(input_knowledge_A) + (0, .5)$) -- ($(input_knowledge_A) + (4, .5)$);
        \node[label={left:Plausible (P)}] (input_knowledge_U) at ($(input_knowledge_A) + (0, 1)$) {};
        \draw[grid] ($(input_knowledge_U) + (0, .5)$) -- ($(input_knowledge_U) + (4, .5)$);
        \node[label={left:Causal (C)}] (input_knowledge_C) at ($(input_knowledge_U) + (0, 1)$) {};
        
        \node[label={below:NP}] (input_param_NO) at (.5,-.5) {};
        \draw[grid] ($(input_param_NO) + (.5, 0)$) -- ($(input_param_NO) + (.5, 3)$);
        \node[label={below:NM}] (input_param_1) at ($(input_param_NO) + (1, 0)$) {};
        \draw[grid] ($(input_param_1) + (.5, 0)$) -- ($(input_param_1) + (.5, 3)$);
        \node[label={below:Pa.}] (input_param_2) at ($(input_param_1) + (1, 0)$) {};
        \draw[grid] ($(input_param_2) + (.5, 0)$) -- ($(input_param_2) + (.5, 3)$);
        \node[label={below:FK}] (input_param_M) at ($(input_param_2) + (1, 0)$) {};

        \draw[very thick] ($(input_knowledge_A) - (0, .5)$) -- ($(input_knowledge_C) + (0, .5)$) -- ($(input_knowledge_C) + (4, .5)$) -- ($(input_param_M) + (.5, 0)$) -- cycle;
        
        \node at ($(input_param_NO)!.5!(input_param_M) + (0, 3.5)$) {\bf Input};

        
        \node[inner sep=0] (time_start) at ($(input_param_M) + (2, 3.5)$) {};
        \node[inner sep=0] (time_end) at ($(input_param_M) + (2, 1.5)$) {};
        \node[inner sep=0] (static_start) at ($(input_param_M) + (2, .5)$) {};
        \node[inner sep=0] (static_end) at ($(input_param_M) + (2, -1.5)$) {};
        
        \draw[very thick, blue] (time_start) -- (time_end);
        \draw[very thick, blue] (static_start) -- (static_end);
        
        \node at ($(time_start) + (0, .3)$) {\bf \textcolor{blue}{Time}};
        \node at ($(static_start) + (0, .3)$) {\bf \textcolor{blue}{Static}};

        
        \node (repr_knowledge_A) at ($(input_knowledge_A) + (7, 0)$) {}; 
        \draw[grid] ($(repr_knowledge_A) + (0, .5)$) -- ($(repr_knowledge_A) + (4, .5)$);
        \node (repr_knowledge_U) at ($(repr_knowledge_A) + (0, 1)$) {};
        \draw[grid] ($(repr_knowledge_U) + (0, .5)$) -- ($(repr_knowledge_U) + (4, .5)$);
        \node (repr_knowledge_C) at ($(repr_knowledge_U) + (0, 1)$) {};
        
        \node[label={below:NP}] (repr_param_NO) at ($(repr_knowledge_A) + (.5, -.5)$) {};
        \draw[grid] ($(repr_param_NO) + (.5, 0)$) -- ($(repr_param_NO) + (.5, 3)$);
        \node[label={below:NM}] (repr_param_1) at ($(repr_param_NO) + (1, 0)$) {};
        \draw[grid] ($(repr_param_1) + (.5, 0)$) -- ($(repr_param_1) + (.5, 3)$);
        \node[label={below:Pa.}] (repr_param_2) at ($(repr_param_1) + (1, 0)$) {};
        \draw[grid] ($(repr_param_2) + (.5, 0)$) -- ($(repr_param_2) + (.5, 3)$);
        \node[label={below:FK}] (repr_param_M) at ($(repr_param_2) + (1, 0)$) {};

        \draw[very thick] ($(repr_knowledge_A) - (0, .5)$) -- ($(repr_knowledge_C) + (0, .5)$) -- ($(repr_knowledge_C) + (4, .5)$) -- ($(repr_param_M) + (.5, 0)$) -- cycle;
        
        \node at ($(repr_param_NO)!.5!(repr_param_M) + (0, 3.5)$) {\bf Representation};

        
        \draw[red, route] ($(input_knowledge_C) + (.5, 0) + 0*(1, 0)$)  node[dot, red] {} to [out=315, in=180] (static_start) node[dot, red] {} to[out=0, in=270]  ($(repr_knowledge_C) + (.75, .25) + 0*(1, 0)$) node[dot, red] {};
        \node[label={[text=red]right:\citep{wen2021causal,kocaoglu2018causalgan,sauer2021counterfactual,depeng,shen2022weakly,van2021decaf}}] at ($(repr_knowledge_C) + (4.5, .25)$) {\textcolor{red}{\dott}};
        
        \draw[DarkTurquoise, route] ($(input_knowledge_A) + (.5, 0) + 1*(1, 0)$)  node[dot, DarkTurquoise] {} to[out=27, in=180] ($(static_start)!.2!(static_end)$) node[dot, DarkTurquoise] {} to[out=0, in=270] ($(repr_knowledge_C) + (.25, -.25) + 1*(1, 0)$) node[dot, DarkTurquoise] {};
        \node[label={[text=DarkTurquoise]right:\citep{cinquini2021}}] at ($(repr_knowledge_C) + (4.5, -.25)$) {\textcolor{DarkTurquoise}{\dott}};
        
        \draw[orange, route] ($(input_knowledge_U) + (.5, 0) + 0*(1, 0)$) node[dot, orange] {} to[out=315, in=180] ($(static_start)!.4!(static_end)$) node[dot, orange] {} to[out=0, in=270] ($(repr_knowledge_C) + (.25, -.25) + 0*(1, 0)$) node[dot, orange] {};
        \node[label={[text=orange]right:\citep{goudet2017causal,Mao_2021_CVPR}}] at ($(repr_knowledge_C) + (6, -.25)$) {\textcolor{orange}{\dott}};
        
        \draw[purple, route] ($(input_knowledge_A) + (.25, .25) + 1*(1, 0)$) node[dot, purple] {} to[out=270, in=180] ($(static_start)!1!(static_end)$) node[dot, purple] {} to[out=0, in=270] ($(repr_knowledge_U) + (.5, 0) + 1*(1, 0)$) node[dot, purple] {};
        \node[label={[text=purple]right:\citep{zhang2022cmgan}}] at ($(repr_knowledge_U) + (4.5, -.25)$) {\textcolor{purple}{\dott}};

        \draw[green, route] ($(input_knowledge_A) + (.25, .25) + 0*(1, 0)$) node[dot, green] {} to[out=270, in=180] ($(static_start)!.6!(static_end)$) node[dot, green] {} to[out=0, in=270] ($(repr_knowledge_U) + (.5, 0) + 0*(1, 0)$) node[dot, green] {};
        \node[label={[text=green]right:\citep{liu2023goggle}}] at ($(repr_knowledge_U) + (4.5, .25)$) {\textcolor{green}{\dott}};
        
        \draw[Violet, route] ($(input_knowledge_A) + (.75, -.25) +  1*(1, 0)$) node[dot, Violet] {} to[out=270, in=180] ($(static_start)!.8!(static_end)$) node[dot, Violet] {} to[out=0, in=270] ($(repr_knowledge_C) + (.75, .25) +  1*(1, 0)$) node[dot, Violet] {};
        \node[label={[text=Violet]right: \citep{li2022gflowcausal}}] at ($(repr_knowledge_C) + (7.5, .25)$) {\textcolor{Violet}{\dott}};

   \end{tikzpicture}
   
   \caption{{\bf Navigating generative modelling.} Here we consider methods that use causality to generate synthetic data. As in \cref{fig:supervised_learning} we discern between methods based on what input they expect (or assume) and what type of representation they learn {\it before} mapping to a label. Here too: Non-parametric (NP), Noise models (NM), Parametric (Pa.), and Fully known (FK).}
   \label{fig:generative}
   \rule{\textwidth}{.5pt}
\end{figure}

\subsubsection{Generative modelling}

Let us now consider \cref{fig:generative} where we map generative models. Note that, like in \cref{fig:bandits} we only consider research that explicitly discusses causality \citep{zhou2023opportunity}. When comparing \cref{fig:bandits,fig:generative}, it should be apparent immediately that a {\it lot} more effort is spent on bandits and RL in CDL. We believe a reason for this is that bandits and RL may provide a more {\it natural} fit with causality given their ability to perform actual interventions in the environment. However, we wish to argue that generative modelling can {\it also} link more tightly with causality. In particular, generative models share an important taks with causal models: both model a distribution.

Most papers in \cref{fig:generative} exploit this parallel. Consider the setting where we are provided a causal model. From \cref{fig:generative} it seems that most researchers focus on this setting in particular. Effectively, given a causal structure, most research aims to incorporate the causal knowledge provided by this structure into the representation space of deep generative models. Any generated data thus respects a given causal graph. Naturally, there are methods that aim to combine structure discovery with generative modelling (providing a {\it structural transition}). However, these methods are equally bound by the same limitations as other causal discovery settings. Specifically because-- unlike bandits and RL --generative modelling typically assumes one observational dataset, rather than access to a live environment in which they can perform interventions.

\end{document}

%% file: math_commands.tex
\usepackage{amsmath,amsfonts,bm}

\def\eqref#1{equation~\ref{#1}}

\def\1{\bm{1}}

\def\eps{{\epsilon}}

\DeclareMathAlphabet{\mathsfit}{\encodingdefault}{\sfdefault}{m}{sl}
\SetMathAlphabet{\mathsfit}{bold}{\encodingdefault}{\sfdefault}{bx}{n}

